\documentclass{article}

\PassOptionsToPackage{numbers, compress}{natbib}

\usepackage[preprint]{neurips_2021}

\usepackage[utf8]{inputenc} 
\usepackage[T1]{fontenc}    
\usepackage{url}            
\usepackage{booktabs}       
\usepackage{amsfonts}       
\usepackage{nicefrac}       
\usepackage{microtype}      
\usepackage{xcolor}         
\usepackage{graphicx}
\usepackage{siunitx,etoolbox}
\robustify\bfseries
\usepackage{booktabs}
\usepackage{multirow}
\def\rot{\rotatebox}
\usepackage{adjustbox}
\usepackage{subfigure}
\usepackage[inline]{enumitem}
\usepackage{hyperref}  
\usepackage{rotating}

\title{The Ikshana Hypothesis of Human Scene Understanding}

\author{%
  Venkata Satya Sai Ajay Daliparthi \\
  Blekinge Insitute of Technology\\
  Karlskrona, Sweden \\
  \texttt{veda18@student.bth.se} \\
}

\begin{document}

\maketitle

\begin{abstract}
In recent years, deep neural networks (DNNs) achieved state-of-the-art performance on several computer vision tasks. However, the one typical drawback of these DNNs is the requirement of massive labeled data. Even though few-shot learning methods address this problem, they often use techniques such as meta-learning and metric-learning on top of the existing methods. In this work, we address this problem from a neuroscience perspective by proposing a hypothesis named Ikshana, which is supported by several findings in neuroscience. Our hypothesis approximates the refining process of conceptual gist in the human brain while understanding a natural scene/image. While our hypothesis holds no particular novelty in neuroscience, it provides a novel perspective for designing DNNs for vision tasks. By following the Ikshana hypothesis, we design a novel neural-inspired CNN architecture named IkshanaNet.
The empirical results demonstrate the effectiveness of our method by outperforming several baselines on the entire and subsets of the Cityscapes and the CamVid semantic segmentation benchmarks.
\end{abstract}

\section{Introduction}
\label{intro}
The human brain can seamlessly perceive diverse perceptual and semantic information regarding the natural scene/image during a glance \cite{friedman1979framing,potter1976short,intraub1981rapid,oliva2001modeling}. The visual scene information perceived during/after a glance refers to the gist (a summary) of the scene/image. The gist includes all the visual information from the low-level (e.g., colors and contours) to the high-level (e.g., shapes and activation). Due to this reason, \cite{oliva2005gist} suggested that the gist can be investigated at both the perceptual and conceptual levels. The structural representation of the image refers to the perceptual gist, and the semantic information of the image refers to the conceptual gist. However, the conceptual gist is more refined and modified than the perceptual gist \cite{oliva2005gist}. Several works \cite{rayner1998eye,biederman1987recognition,oliva2001modeling,oliva2000diagnostic,henderson2003human,evans2005perception,fei2007we} in neuroscience have addressed the fundamental question, i.e.,``how does the human brain performs several visual tasks?" by investigating through conceptual and perceptual gist. They conducted several experiments and proposed various theories to explain how modeling of the scene occurs in the human brain. However, there was no general principle that explains the functioning of the human brain. Even though there is a general principle, we expect that to be different from human-to-human. Depending on the situation and the environment, the human brain can seamlessly grasp the information by recognizing the objects and observing their structure. On the other hand, for a computer to do the same is the fundamental goal of the computer vision field. \newline
In recent years, deep learning methods have shown a significant improvement over traditional handcrafted techniques on several computer vision tasks. Though these deep neural networks (DNNs) achieved state-of-the-art performance in many cases, the one major drawback is the requirement of massive labeled data. The collection of a huge amount of labeled data is an expensive and time taking process. Even though these DNNs are said to be inspired by the functioning of the human brain, is this how the human brain learns to perform any visual task? \textbf{NO}. Because the human brain does not require massive labeled data to perform any visual task, and it can perform with few data samples. However, we cannot observe a similar phenomenon in the case of many DNNs. \newline
\textbf{Semantic segmentation} is the task of assigning a class label to every pixel in the given image, which has applications in various fields such as medical, autonomous driving, robotic navigation, localization, and scene understanding. The prominent work FCN \cite{long2015fully} adopted the image-classification networks \cite{krizhevsky25hinton,simonyan2014very,7298594} for semantic segmentation. Later on, several works \cite{ronneberger2015u,badrinarayanan2017segnet,zhao2017pyramid,chen2018encoder,SunXLW19,10.1007/978-3-030-58539-6_11,takikawa2019gated,yuan2020segfix} improved the FCN \cite{long2015fully} architecture, and proven to be successful in diverse semantic segmentation benchmarks \cite{BrostowSFC:ECCV08,cordts2016cityscapes,8100027}. However, these methods mainly focus on achieving state-of-the-art performance by using the entire and additional datasets \cite{deng2009imagenet} (for pre-training). Due to this reason,
even though various methods \cite{chen2018encoder,SunXLW19} outperformed U-Net \cite{ronneberger2015u} in terms of accuracy and computational complexity, the U-Net \cite{ronneberger2015u} architecture is still exploited in several medical image segmentation methods due to its ability to perform with few data samples \cite{siddique2020u}.
Although 
several few-shot semantic segmentation (FSS) methods are introduced to address this problem, they often use techniques such as meta-learning \cite{xu2016deep,dong2018few,RakellySDEL18,2009-06680,TianWQWSG20} and metric learning \cite{BMVC2017_167,9108530,Wang_2019_ICCV,Zhang_2019_CVPR,WangZHYCZ20,WangZHYCZ20,Zhang_2019_ICCV,YangLLJY20} on top of the existing  architectures. \newline
Unlike FSS methods, we tackle the formerly mentioned drawback of the DNNs, i.e., the requirement of massive labeled data, from a neuroscience perspective. In this work, we propose a hypothesis of human scene understanding mechanism named Ikshana. The idea is that, ``to understand the conceptual gist of a given image; humans look at the image multiple times recurrently at different scales''. Following the Ikshana hypothesis, we propose a novel neural-inspired CNN architecture named IkshanaNet, a multi-scale architecture that learns representations at full image resolution. 
In contrast to the existing CNN architectures that pass the input image only to the initial layer (stem module), our method feeds the input image to every module in the network and to the best of our knowledge, this is the first work to propose the same.\newline
To evaluate the performance of IkshanaNet,  we conduct extensive experiments on the entire and subsets of the Cityscapes and Camvid benchmarks. Moreover, we conduct multiple ablation studies to verify the effect of image scales in IkshanaNet.
The empirical results illustrate that our method outperforms several baselines on the entire and few data samples. Furthermore, the ablation studies shows the importance of multi-scale information to achieve considerable performance.
We hope that our hypothesis sparks future research in neural network architectures for vision tasks.
\section{Related work}
\label{related work}
In \textbf{Neurological} terms, all the low-level and high-level computer vision tasks come under a single term called human scene understanding. A scene is a view of a real-world environment that contains multiple surfaces and objects organized in a meaningful way. In neuroscience, the perceptual gist is more investigated compared to the conceptual gist. The early works on the conceptual gist \cite{potter1975meaning,intraub1981rapid} explained that a typical scene fixation of $275$ to $300$ $ms$ is often sufficient to understand the gist of the image. Several works on the perceptual gist \cite{biederman1987recognition,rayner1998eye,schyns1994blobs,oliva1997coarse,oliva2000diagnostic,oliva2001modeling,henderson2003human,evans2005perception,fei2007we} provided insight into how the modelling of the scene occurs in the human brain through perceiving boundaries, blobs, scales, texture, contours, openness, depth, and so on. The information perceived through the perpetual gist is refined and extracted into the conceptual gist (the semantic meaning) during the cognitive process. Thus, the conceptual gist is highly dependent upon the perceptual gist.
In many cases \cite{deng2009imagenet,cordts2016cityscapes,8100027}, we do not explicitly encode the perceptual process in DNNs, and the CNN learns various representations regarding the image during the training process. Thus, our hypothesis focuses on the conceptual gist rather than the perceptual gist. \newline
\textbf{Neural networks} exist from a long time  \cite{mcculloch1943logical,rumelhart1986learning,rosenblatt1958perceptron} and some prominent works  \cite{deng2009imagenet,krizhevsky25hinton,lin2013network,simonyan2014very,7298594,he2016deep,44903,8099678} made them popular during recent years. In our work, we use the convolutional neural network (CNN) architecture  \cite{lecun1998gradient,21701} to learn representations from the images, which itself is inspired by \cite{hubel1962receptive,fukushima1983neocognitron}. The architecture of IkshanaNet is inspired by \cite{simonyan2014very,huang2017densely} and related to \cite{huang2018multiscale,liao2016bridging}.\newline
The first seminal work on \textbf{Semantic segmentation (SS)} using deep learning is the fully convolutional networks (FCN) \cite{long2015fully}. Later on, many semantic segmentation networks followed the FCN \cite{long2015fully} architecture. The total prominent works on deep learning-based semantic segmentation methods can be roughly classified into five categories.  They are (i) Encoder-decoder based methods (DeconvNet \cite{noh2015learning}, SegNet \cite{badrinarayanan2017segnet}, U-Net \cite{ronneberger2015u}, RefineNet \cite{Lin:2017:RefineNet,lin2019refinenet}, FC-DenseNet \cite{fc-densenet}, and GFR-Net \cite{islamsal18}), (ii) Regional proposal methods (MaskRCNN \cite{he2017mask}, FPN \cite{lin2017feature}, and PANet \cite{liu2018path}), (iii) Increased resolution of feature map methods (DeepLab series \cite{ChenPKMY14,7913730,ChenPSA17,chen2018encoder}, PSPNet \cite{zhao2017pyramid}, DenseASPP \cite{8578486}, and HRNet \cite{SunXLW19}), (iv) Context information methods (ParseNet \cite{LiuRB15}, ATS \cite{7780765}, DANet \cite{fu2019dual}, OCNet \cite{abs-1809-00916}, OCR \cite{10.1007/978-3-030-58539-6_11}, EncNet \cite{Zhang_2018_CVPR}, Non-local \cite{8578911}, ZigZagNet \cite{8954370}, ACFNet \cite{9010415}, CoCurNet \cite{8954430}, GLAD \cite{xi}, and HANet \cite{choi2020cars}) (v) Boundary refinement methods ( \cite{Bertasius_2015_ICCV,CB2016Semantic,Ding_2019_ICCV,MarinHVCTYB19}, Gated-SCNN \cite{takikawa2019gated}, and SegFix \cite{yuan2020segfix}).
The IkshanaNet uses the dilated convolutions, interpolation of feature maps, and skip connections from different layers in the network. Therefore, our work is related to the formerly mentioned encoder-decoder and increased resolution of feature map methods.\newline
\textbf{Few-shot segmentation (FSS)} methods \cite{xu2016deep,dong2018few,RakellySDEL18,2009-06680,TianWQWSG20,BMVC2017_167,9108530,Wang_2019_ICCV,Zhang_2019_CVPR,WangZHYCZ20,Zhang_2019_ICCV,YangLLJY20,GFSS,meta-seg,Dyanamic-Extension-GFSS} are introduced to handle limited training data. They use meta-learning (knowledge distillation), metric-learning (similarity learning), and a combination of both the techniques on top of FCN \cite{long2015fully} based architectures, which often involve multistage training. The metric-learning techniques can be further classified into the prototypical feature learning \cite{dong2018few,priorFSS,Wang_2019_ICCV,Zhang_2019_CVPR,9108530,ASGNet} and the affinity learning \cite{WangZHYCZ20,Zhang_2019_ICCV,YangWZCYOZ20} techniques. Unlike general SS methods, FSS methods are evaluated on different benchmarks and handle novel class categories during testing.
Since the IkshanaNet does not use any of the formerly mentioned FSS techniques and only handles the classes seen in the training data, our method is more closely related to the general SS methods than the FSS methods.
\section{Method}
\label{method}
\subsection{Ikshana (the eye) hypothesis}
\label{3.1}
\begin{figure}[ht]
\begin{center}
   \includegraphics[width=0.8\linewidth]{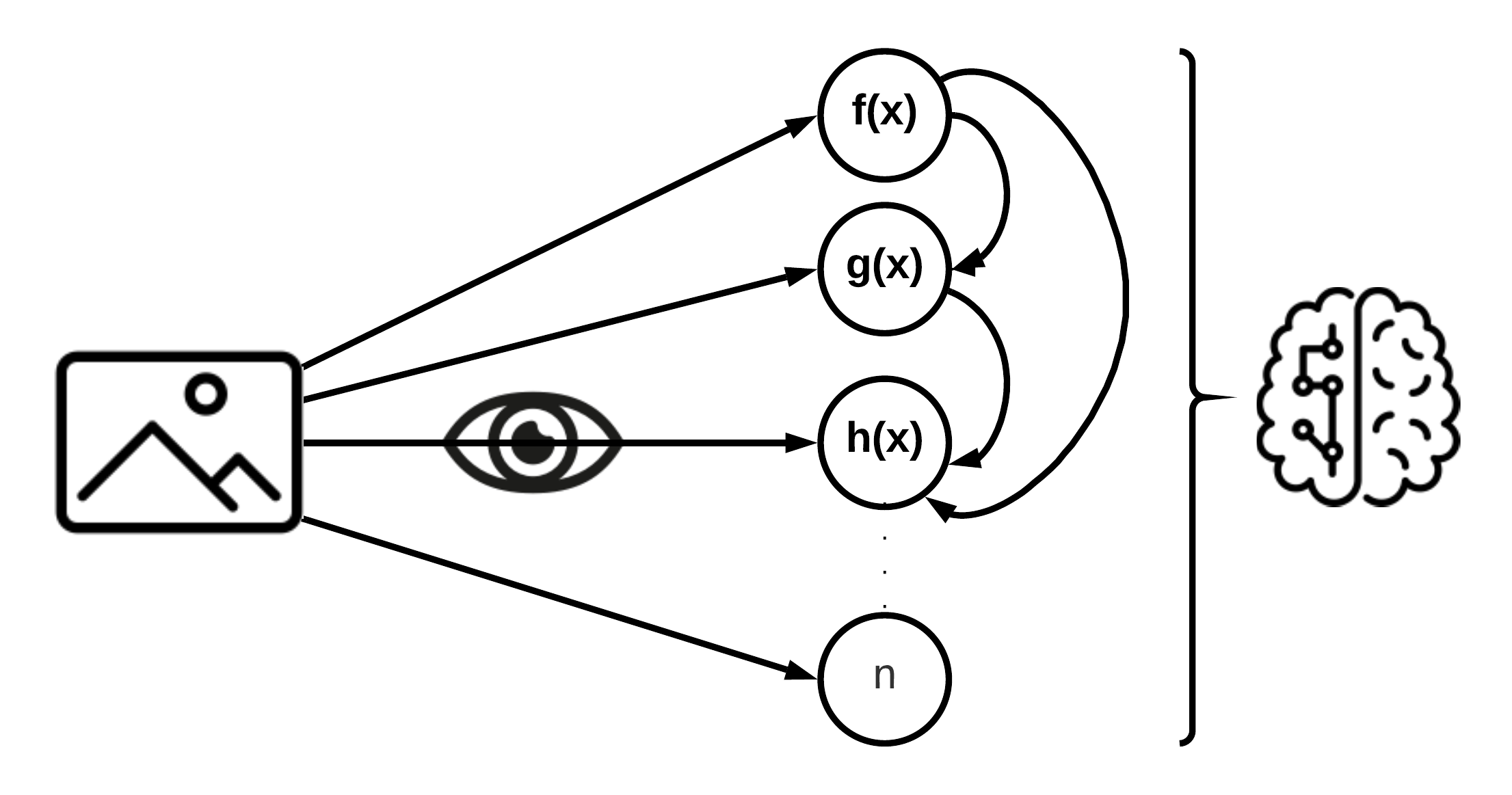}
\end{center}
   \caption{The Ikshana hypothesis at single scale}
\label{fig:ikshana}
\end{figure}

In her prominent work \cite{potter1975meaning}, professor Mary C. Potter found that an average human can understand the gist of the image between the time interval of $125$ to $300$ $ms$. Furthermore, through several works \cite{intraub1981rapid,rayner1998eye,schyns1994blobs,oliva2000diagnostic,oliva2001modeling,henderson2003human,evans2005perception,fei2007we} in neuroscience, it is evident that humans understand the gist of the image in a certain time interval. During that time interval, the Ikshana hypothesis approximates the functioning of the human brain.
The Ikshana hypothesis states that \textbf{``To  understand the conceptual gist of a given image, humans look at the image multiple times recurrently, at different scales.''} The word Ikshana is derived from the Sanskrit language, which has many synonyms such as the eye, sight, look, and so on. \newline
We present an example to explain the Ikshana hypothesis in Figure \ref{fig:ikshana}, where there is an image ($x$) on the left side and the human brain mechanism on the right side.
According to the Ikshana hypothesis, for a human to understand the conceptual gist of the given image, the following process occurs in the human brain: \newline
At a time step ($t$), during the first glance ($\Phi_{1}$), the brain learns the first representation ($f(x)$) from the image ($x$) and stores that representation in the memory ($M$), as shown in the equation \ref{eq1}.
\begin{equation} \label{eq1}  
 f(x) = \Phi_{1}(x)    ; \quad  M = f(x) 
\end{equation}
At a time step ($t+1$), during the second glance ($\Phi_{2}$), the brain holds the first representation ($f(x)$) in the memory and learns the second representation ($g(x)$) from the image and the first representation ($x,f(x)$). Then the brain stores the representation ($g(x)$) along with ($f(x)$) in the memory ($M$), as shown in the equation \ref{eq2}.
\begin{equation} \label{eq2}  
 g(x) = \Phi_{2}(x,f(x))   ;\quad   M = f(x),g(x)
\end{equation}
At a time step ($t+2$), during the third glance ($\Phi_{3}$), the brain holds the first and the second representations ($f(x)$, $g(x)$) in the memory and learns the third representation ($h(x)$) from the image and the previous representations $(x,f(x),g(x))$. Then the brain stores the representation ($h(x)$) along with ($f(x)$, $g(x)$) in the memory ($M$), as shown in the equation \ref{eq3}.  
\begin{equation} \label{eq3}  
 h(x) = \Phi_{3}(x,f(x),g(x))   ;  \quad M = f(x),g(x),h(x)
\end{equation}
From \ref{eq1}, \ref{eq2}, and \ref{eq3}, this kind of recurrent process occurs at ($t+n$) times at a single image scale. Depending upon the given task ($T$), by combing all the information stored in the memory until the ($t+n$)th time step, the brain understands the conceptual gist ($Y_{1}$) of the image at a single scale, as shown in the equation \ref{eq4}. 
\begin{equation} \label{eq4}  
 Y_{1} = T(f(x), g(x), h(x)...........n(x))
\end{equation}
This process occurs at $N$ different scales and generates $N$ different outputs ($Y_{1}$, $Y_{2}$, $Y_{3}$, ...., $Y_{n}$). By considering all the outputs, the brain selects some of those representations and forgets the remaining representations. In this way, the brain learns ($\Delta$) the final output ($Y$) of the given visual task ($T$), as shown in the equation \ref{eq5}.
\begin{equation} \label{eq5}  
 Y = \Delta ( Y_{1},Y_{2},Y_{3},....Y_{N})
\end{equation}
From the equations \ref{eq1}, \ref{eq2}, \ref{eq3}, \ref{eq4}, and \ref{eq5}, this is how Ikshana hypothesis approximates the functioning of the human brain, while human understands the conceptual gist of the image.
The time required (or the number of glances required) by an average human to understand the gist of the image may depend upon several factors such as the given task, age, intelligence, memory, and so on. 

\noindent The existing CNN architectures such as VGG \cite{simonyan2014very}, Resnet \cite{he2016deep}, DenseNet \cite{huang2017densely}, and so on learns a representation (say $f(x)$) with $32/64$ filters from the input image and learns further representations on top of the $f(x)$ until the network achieves adequate performance. In contrast, the network designed by following the Ikshana hypothesis learns representations from the input image and previous outputs at each glance/layer.


\subsection{IkshanaNet-main}
\label{3.2}
\begin{figure}[ht]
\begin{center}
   \includegraphics[width=0.95\linewidth]{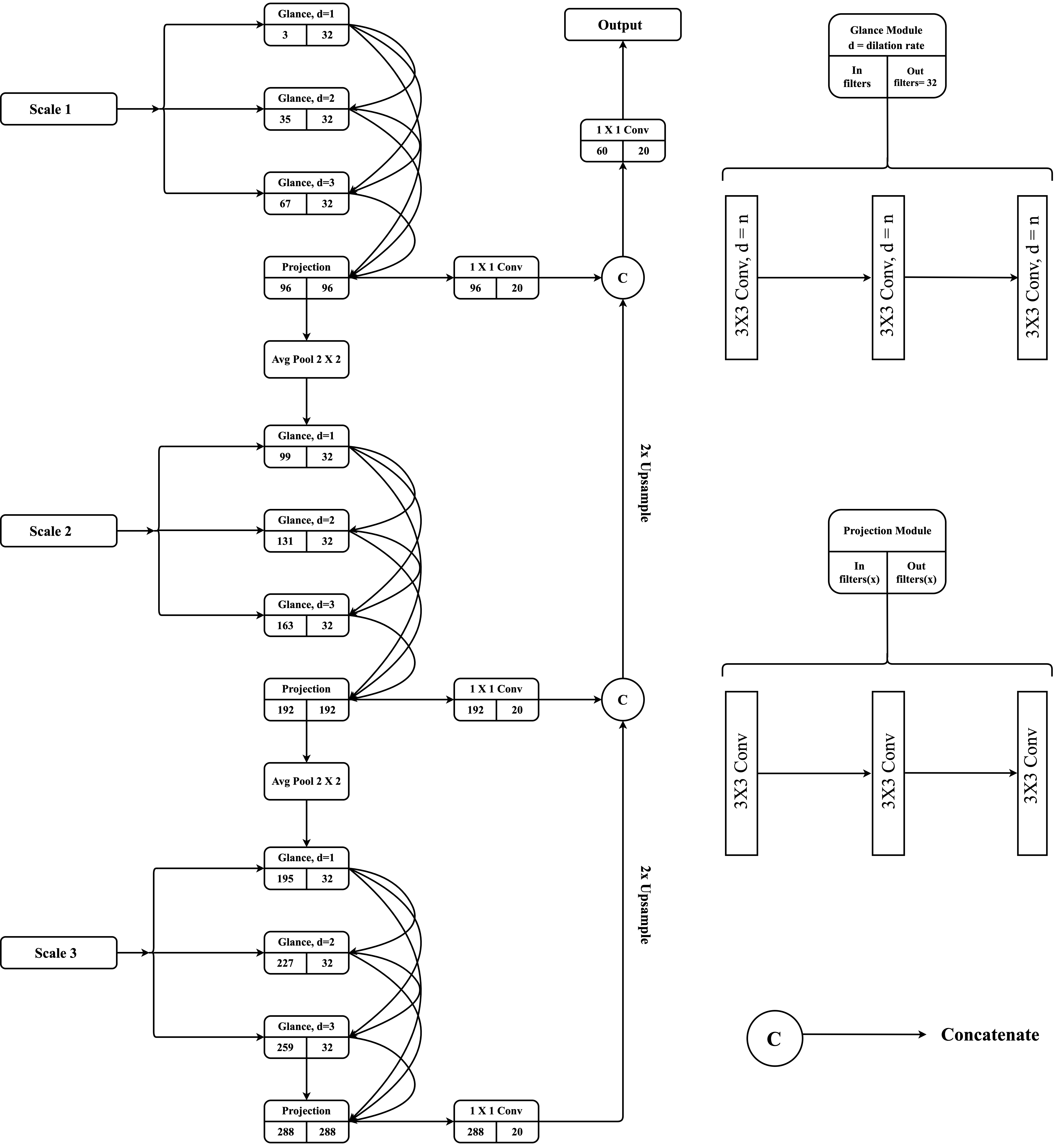}
\end{center}
   \caption{IkshanaNet-main architecture}
\label{fig:IkshanaNet}
\end{figure}

\noindent In this section, we introduce a novel neural-inspired encoder-decoder CNN architecture named IkshanaNet, designed by following the Ikshana hypothesis. Humans can look at the image and seamlessly learn various useful representations regarding it \cite{friedman1979framing,potter1976short,intraub1981rapid,oliva2001modeling}. On the other hand, for a computer to do the same, we use the convolutional neural network \cite{hubel1962receptive,fukushima1983neocognitron,lecun1998gradient} architecture to learn representations. The IkshanaNet architecture uses three image scales and consists of $4$M parameters. The entire architecture is made of three building blocks, and they are: ($1$) the glance module, ($2$) the projection module, and ($3$) a $1$x$1$ convolutional layer, as illustrated in Figure \ref{fig:IkshanaNet}.\newline
The \textbf{glance module} consists of three $3$x$3$ convolutional layers (with the same dilation rates), and we use it to learn representations from the given image (or a feature map). The number of input filters passed into the glance module varies several times in the architecture; however, it always returns a feature map with $32$ filters.
The \textbf{projection module} consists of three $3$x$3$ convolutional layers, and we use it to refine the representations learned from the glance modules. The input and output filters are always the same for the projection module.
We use the \textbf{$1$x$1$ convolution layers} to reduce the number of filters in a given future map. Except for the last $1$x$1$ convolutional layer that returns the final output, every  convolutional layer in the architecture is followed by a batch normalization \cite{ioffe2015batch} and a ReLU \cite{nair2010rectified} activation layer. \newline 
In \textbf{the encoder} part, the IkshanaNet learns representations at three image scales. At scale $1$, we pass the input image through a glance module with a dilation rate (d$=1$), which returns a feature map with $32$ filters. Then we concatenate the input image with the previously learned feature map $(32+3=35)$. The concatenation of the input image with the feature map is essential to ensure that we are learning representations from the input image. Then we pass the feature map through another glance module with a dilation rate (d$=2$) and concatenate the resulting feature map with the feature maps from the preceding layers $(32+32+3=67)$. We pass the resulting feature map through another glance module with a dilation rate (d$=3$), which takes in $67$ filters and returns $32$ filters. Again, we concatenate the resulting feature map with feature maps from the preceding layers $(32+32+32+3=99)$. At this point, we remove the input image from the feature map through tensor slicing $(99-3=96)$, and the resulting feature map consists of $(32+32+32=96)$ filters learned from three glances modules. In this way, the network followed the Ikshana hypothesis had three glances recurrently at the full resolution. Then we pass the feature map through a projection module to refine the representations $(96=96)$. Here, we pass the refined feature map through a $1$x$1$ convolutional layer that reduces $96$ filters into $20$ filters and name it the side one output ($Y_{1}$). Simultaneously, we pass the feature map through an average pooling layer, which reduces the size of the feature map by a factor of two.\newline
At scale $2$, we down-sample the input image by a factor of two and concatenate with the pooled feature map from the scale $1$ $(96+3=99)$. We pass the resulting feature map with $99$ filters through three glance modules with different dilation rates (d=$1$, $2$, $3$) and concatenate all the outputs as follows $(99+32+32+32=195)$. Then we remove the image from the feature map $(195-3=192)$ and pass it through a  projection module to refine the representations $(192=192)$. Then we pass the refined feature map through a $1$x$1$ convolutional layer that reduces $192$ filters into $20$ filters and name it the side two output ($Y_{2}$). Then, we pass the refined feature map through an average pooling layer that reduces the size by a factor of two.\newline
At scale $3$, we down-sample the input image by a factor of four and concatenate with the pooled feature map from the scale $2$ $(192+3=195)$. Here, we follow the same process $(195+32+32+32=291);(291-3=288);(288==288)$ as the scale $2$ part, which returns a feature map with $20$ filters, and name it the side three output ($Y_{3}$). \newline
In \textbf{the decoder} part, we bi-linearly interpolate the outputs from two scales ($Y_{2}$ and $Y_{3}$) to match with the output of scale $1$ $Y_{1}$, i.e., the input image size. Then we concatenate all the three outputs $(20+20+20=60)$ and pass it through a $1$x$1$ convolutional layer, which returns a feature map with $20$ filters, that is the final output of the network. $[Y=\Delta(Y_{1}, Y_{2},Y_{3})]$ \newline
\textbf{Depth architectures :}
Here, we introduce three variants of the IkshanaNet named IkshanaNet-$3$G, IkshanaNet-$6$G, and IkshanaNet-$12$G.  If we remove the projection layers in IkshanaNet-main, then it will remain with three scales and three glances at each scale; it is IkshanaNet-$3$G (which consists of $514$K parameters). If we increase the number of glances per scale, from three to six, then it is IkshanaNet-$6$G (which consists of $1.8$M parameters), and from three to twelve, then it is IkshanaNet-$12$G (which consists of $6.5$M parameters).\newline
\textbf{Multi-scale architectures :}
Here, we introduce three variants of IkshanaNet named IkshanaNet $1$S-$6$G, $2$S-$3$G, and $3$S-$2$G. In IkshanaNet $1$S-$6$G, there are no pooling layers and contain six glances at full-scale resolution (which consists of $257$K parameters). In IkshaNet $2$S-$3$G, there are two scales and three glances at each scale (which consists of $259$K parameters). In IkshanaNet $3$S-$2$G, there are three scales and two glances at each scale (which consists of $260$K parameters).

\section{Experiments}
\label{4}
\subsection{Experimental setup}
\label{experimental-setup}

\noindent \textbf{GPU:} $1$ X NVIDIA Tesla T-$4$ ($16$ GB VRAM)\newline
\textbf{Framework :} PyTorch $1.8$ \cite{NEURIPS2019_9015}\newline
\textbf{Epochs :} $180$ ;
\textbf{Batch size :} $2$\newline
\textbf{Criterion :} Pixel-wise cross-entropy loss \newline
\textbf{Learning rate scheduler :} ReduceLROnPlateau (decrease factor = $0.5$ and patience = $20$ epochs) with an initial learning rate of $1e-06$. \newline
\textbf{Optimizer :} Stochastic gradient descent \cite{robbins1951stochastic} with Nesterov momentum \cite{nesterov1983method} \footnote{ For all the baselines, we use the Nesterov momentum  of $0.9$ for the SGD \cite{robbins1951stochastic} optimizer by following \cite{chen2018encoder,huang2017densely,he2016deep,mobileNetV2,TanL19,RegNet}. For the IkshanaNet and its variants, we use the Nesterov momentum of $0.7$ for the SGD \cite{robbins1951stochastic} optimizer by tuning with several values such as $0.5, 0.6, 0.7, 0.8$, and $0.9$, i.e., the only hyper-parameter tuning step in this work. In our preliminary experiments, we observe that the training of IkshanaNet is unstable with $0.9$ momentum. We hypothesize that this phenomenon is due to the small size of IkshanaNet compared to baseline networks} \newline
\textbf{Random seed :} To ensure that data splits are reproducible, we set the random seed $42$ in the function torch.utils.data.random-split. \newline
\textbf{Pre-processing} We normalize all the images with mean and standard deviation values of ImageNet \cite{deng2009imagenet} dataset. We did not use any data augmentation techniques.\newline
\textbf{Baselines:} We use the open-source implementations for networks DeepLabV3+ (ResNet-101) \cite{Resnet-101}, DeepLabV3 (DenseNet-161) \cite{Densenet-161}, HRNet-V2 \cite{HRNet-V2},  and U-Net \cite{U-Net}. We import DeeplabV3+ with encoder networks such as ResNet \cite{he2016deep}, MobileNet-V2 \cite{mobileNetV2}, ResNext \cite{Resnext}, EfficientNet \cite{TanL19}, and RegNet \cite{RegNet} from the segmentation models library \cite{Yakubovskiy:2019}.

\subsection{Experiments on Cityscapes}
\label{4.2}
The Cityscapes \cite{cordts2016cityscapes} semantic segmentation dataset consists of $5,000$ finely annotated high-quality images, which are further divided into $2,975/500/1,525$ images for training, validation, and testing.  During the evaluation, only $19$ classes are considered out of the $35$ classes. Therefore, by using the cityscapes-scripts, we convert the $35$ classes into $20$ classes (including background). We resize all the images from the resolution of $1024$x$2048$ to $512$x$1024$. 
\subsubsection{Baseline experiments :}
\label{4.2.1}
Here, we use the networks DeeplabV3+ (ResNet-101 \cite{he2016deep}), DeeplabV3 (DenseNet-161 \cite{huang2017densely}), HRNet-V2 \cite{SunZJCXLMWLW19}, and U-Net \cite{ronneberger2015u} as the baselines \footnote{ For the baselines, ResNet-101 \cite{he2016deep}, DenseNet-161 \cite{huang2017densely}, and HRNet-V2 \cite{SunZJCXLMWLW19}, we use the ImageNet \cite{deng2009imagenet} pre-trained weights. Because in the existing literature, the architectures \cite{chen2018encoder,zhao2017pyramid,10.1007/978-3-030-58539-6_11,SunZJCXLMWLW19} used an ImageNet pertained network as a feature extractor and reported the results by using pre-trained weights only. However, in the case of IkshanaNet and U-Net \cite{ronneberger2015u} no pre-training is done. Since this work addresses the requirement of massive data, this provides strong motivation against pre-training.} to compare with IkshanaNet-main.\newline
We train all the networks on the entire dataset $T_{2975}$ and provide the mean class IoU results evaluated on the validation-set in Table \ref{table-1}, where we observe the following: \newline
(i) U-Net \cite{ronneberger2015u} ($49.3$) shown top performance within the baseline networks followed by  HRNet-V2 \cite{SunZJCXLMWLW19} ($48.0$).\newline
(ii) IkshanaNet outperformed U-Net by $5.2$ \% and HRNet-V2 \cite{SunZJCXLMWLW19} by $6.5$ \%.\newline
(iii) IkshanaNet outperformed baselines by a huge margin in classes such as fence, pole, traffic light, traffic sign, rider, bus, motorcycle, and bicycle.\newline
(iv) Even though U-Net \cite{ronneberger2015u} and IkshanaNet learn representations at full-scale resolution before reducing the spatial resolution, the IkshanaNet still outperforms U-Net \cite{ronneberger2015u} in the formerly mentioned classes.
\begin{table}[ht]
\sisetup{detect-weight=true}
\caption{Class-wise IoU results of the Cityscapes baseline experiments}
\begin{center}
\begin{adjustbox}{width=1\textwidth}
\begin{tabular}{lSSSSSSSSSSSSSSSSSSSS}
 \toprule
  {Method} & \rot{90}{road} & \rot{90}{sidewalk} &\rot{90}{building}  & \rot{90}{wall} &\rot{90}{fence}  & \rot{90}{pole} &\rot{90}{traffic light}  & \rot{90}{traffic sign} &\rot{90}{vegetation}  & \rot{90}{terrain} &\rot{90}{sky}  & \rot{90}{person} &\rot{90}{rider}  & \rot{90}{car} &\rot{90}{truck}  & \rot{90}{bus} &\rot{90}{train}  & \rot{90}{motorcycle} &\rot{90}{bicycle}  & \rot{90}{Average}      \\
  \midrule
ResNet101 \cite{he2016deep} &  95.0 & 66.1 & 81.9& 15.0 & 13.5 & 26.7 & 20.7 & 29.5 & 86.7 & 55.4 & 89.3 & 48.5 & 6.3 & 85.5 & 6.8 & 26.1 & 19.0 & 9.8 & 32.0 & 42.8\\
DenseNet161 \cite{huang2017densely} & 94.8 & 64.5 & 81.3& 20.1 & 13.0 & 15.8 & 15.6 & 28.7 & 84.6 & \bfseries 58.7 & 86.1 & 44.1 & 0.6 & 84.7 &  17.0 & 19.7 & \bfseries 23.1 & 4.3 & 31.4 & 41.5\\
HRNet-V2 \cite{SunZJCXLMWLW19}&  94.9 & 68.6 & 84.2& 24.0 & 24.5 & 39.0 & 23.2 & 42.3 & 86.9 & 51.5 & 90.2 & 55.6 & 15.3 & 86.1 & \bfseries 19.9 & 36.1 & 21.2 & 2.2 & 46.1 & 48.0\\
U-Net \cite{ronneberger2015u} &  94.9 & 69.4 & 85.3 & \bfseries 27.3 & 28.7 & 41.0 & 32.2 & 49.0 & 88.6 & 46.3 & 90.4 & 59.1 & 14.5 & 86.5 & 12.4 & 28.4 & 15.5 & 10.9 & 55.6 & 49.3\\
IkshanaNet-Main  &  \bfseries 95.6 & \bfseries 72.8 & \bfseries 85.9 & 22.6 & \bfseries 35.3 & \bfseries 49.6 & \bfseries 47.0 & \bfseries 60.7 & \bfseries 89.2 & 48.9 & \bfseries 91.6 & \bfseries 63.3 & \bfseries 28.8 & \bfseries 87.1 & 18.4 & \bfseries 40.3 & 21.8 & \bfseries 16.5 & \bfseries 60.8 & \bfseries 54.5\\
\bottomrule
\end{tabular}
\end{adjustbox}
\end{center}
\label{table-1}
\end{table}

\subsubsection{Data ablation study :}
\label{4.2.2}
While trained on few data samples, the network size might strongly influence the performance. The networks ResNet-101 \cite{he2016deep} ($59.3$ M), DenseNet-161 \cite{huang2017densely}) ($43.2$ M), HRNet-V2 \cite{SunZJCXLMWLW19} ($65.9$), and U-Net \cite{ronneberger2015u} ($31.0$) consists more number of parameters compared to IkshanaNet-main ($4$ M). To make it a fair comparison, we include DeeplabV3+ \cite{chen2018encoder} with several light-weight encoder networks (such as ResNet-18 \cite{he2016deep}, MobileNet-V2 \cite{mobileNetV2}, EfficientNet-b1 \cite{TanL19}, and RegNetY-08 \cite{RegNet}) along with the networks from the baseline experiments.\newline
Here, we conduct a data ablation study on five different subsets of the training data, $T_{1487}$, $T_{743}$, $T_{371}$, $T_{185}$, and $T_{92}$ (suffix number represents the number of training samples in the subset)  by using the same validation set ($500$ images). \newline
In table \ref{Table2}, we provide the mean class IoU results evaluated on the validation set,  the average M.IoU score, the number of parameters (in million)), and the GFLOPs \cite{GFLOPS} (calculated with an input resolution of $1$x$512$x$1024$x$3$ ). \newline
From table \ref{Table2}, we observe the following: \newline
(i) U-Net \cite{ronneberger2015u} ($T_{avg}$- 32.2) achieves top average performance within the baselines\newline (ii) Even though U-Net \cite{ronneberger2015u} consists of $31$M parameters, it still managed to outperform its lightweight counterparts.\newline
(iii) IkshanaNet outperformed all other baselines in the M.IoU score and the average M.IoU score in all five subsets.\newline
(iv) IkshanaNet consists of fewer parameters, and EfficientNet-b1 \cite{TanL19} consists of fewer GFLOPs than other networks.

\begin{table}[ht]
\sisetup{detect-weight=true}
\caption{Cityscapes data ablation experiments evaluated on the validation set}
\begin{center}
\begin{adjustbox}{width=1\textwidth}
  \begin{tabular}{lSSSSSSSS}
    \toprule
     {Backbone} &
     {$T_{1487}$} & {$T_{743}$} & {$T_{371}$} & {$T_{185}$} & {$T_{92}$} & {\textbf{$T_{avg}$}} & {Param(M)} &{GFLOPs}  \\
      \midrule
    ResNet-18  \cite{he2016deep} & 42.6 & 35.6 & 27.9 & 22.4 & 21.0 & 29.9 & 12.3 & 36.8 \\
    MobileNet-V2 \cite{mobileNetV2} & 38.5 & 32.2 & 30.6 & 22.5 & 19.2 & 28.6 &  4.4& 12.3 \\
    EfficientNet-b1 \cite{TanL19} & 37.8 & 32.5 & 26.9 & 24.6 & 19.8 & 28.3 & 7.4 & \bfseries 4.6 \\
    RegNetY-08 \cite{RegNet} &  28.5& 31.9 & 29.4 & 27.4 & 22.1 & 27.9  & 7.0 & 17.2 \\
    ResNet-101 \cite{he2016deep} & 29.3 & 28.8 & 28.6 & 21.6 & 19.4 & 25.5 & 59.3 & 177.8 \\
    DenseNet-161 \cite{huang2017densely} & 33.3 & 30.1 & 26.0 & 24.9 & 20.8 & 27.0 & 43.2 & 129.4 \\
    HRNet-V2 \cite{SunZJCXLMWLW19} & 27.8 & 18.8 & 23.3 & 18.3 & 15.4 & 20.7 & 65.9 & 187.8 \\
    U-Net\cite{ronneberger2015u} & 42.8 & 34.2 & 30.2 & 27.8 & 25.0 & 32.0 & 31.0 & 387.1 \\
    IkshanaNet-Main &  \bfseries 43.4 & \bfseries 40.2 & \bfseries 31.7 & \bfseries 29.9 & \bfseries 25.8 &  \bfseries 34.2  & \bfseries 4.0  & 413.3 \\

    \bottomrule 
  \end{tabular}
\end{adjustbox}
\end{center}

 \label{Table2}
\end{table}

\subsubsection{Multi-scale ablation study :}
\label{4.2.3}
 In Section \ref{3.1}, the Ikshana hypothesis stated that ``humans often require multi-scale information to understand the gist of an image''. Therefore, to verify the requirement of multi-scale information, we conduct a multi-scale ablation study.\newline
Here, we train three different variants of IkshanaNet, such as the $1$S-$6$G, $2$S-$3$G, and $3$S-$2$G (explained in Section \ref{3.2}) on the five different subsets of the training data (same as Section \ref{4.2.2}).
In table \ref{Table-3}, we provide the results of the multi-scale ablation study evaluated on the validation set.\newline
From table \ref{Table-3}, we observe that:\newline
(i) IkshanaNet-3S-2G network outperforms other networks in the M.IoU score, the average M.IoU score, and requires fewer GFLOPs, while requiring the same number of parameters.\newline
(ii) The multi-scale information improved the performance and decreased the computational complexity (GFLOPs) of the network and vice-versa.\newline
(iii) From Table \ref{Table2} and Table\ref{Table-3}, we observe that IkshanaNet $3$S-$2$G network (with only $260$K parameters) outperforms all the baselines in the data ablation study by occupying approximately $10$x few GFLOPs and $15$x few parameters than IkshanaNet-main.\newline
The above observations suggest that, the multi-scale architectures can achieve superior performance than an isometric architecture.

\begin{table}
\caption{Cityscapes multi-scale ablation experiments evaluated on the validation set}
\sisetup{detect-weight=true}
 \centering
  \begin{adjustbox}{width=1\textwidth}
  \begin{tabular}{lSSSSSSS[table-format=1.2]S}
    \toprule
     {Backbone} &
     {$T_{1487}$} & {$T_{743}$} & {$T_{371}$} & {$T_{185}$} & {$T_{92}$} & {\textbf{$T_{avg}$}} & {Param(M)} &{GFLOPs}  \\
      \midrule
    1S-6Glances & 29.2 & 24.9 & 23.3 & 20.2  &18.1  & 23.1 &   0.26 & 136.0 \\
    2S-3Glances & 37.3 & 34.9 & 33.2 & 25.7 & 24.0 & 31.0 & 0.26  & 70.0 \\
    3S-2Glances & \bfseries 43.5 & \bfseries 36.9 & \bfseries 34.4 & \bfseries 27.5 & \bfseries 26.5 & \bfseries 33.8& 0.26 & \bfseries 42.4 \\
    \bottomrule
  \end{tabular}
  \end{adjustbox}
  \label{Table-3}
\end{table}
\subsection{Experiments on Camvid}
\label{4.3}
The Cambridge-driving labeled video dataset \cite{BrostowSFC:ECCV08} for semantic segmentation consists of $700$ images, which are further divided into $367$ training, $101$ validation, and $233$ testing sets. We convert the $32$ classes to $12$ classes (including background) by following \cite{badrinarayanan2017segnet,segnet} and resize the images from the resolution of $720$x$960$ to $368$x$480$. 

\begin{table}[ht]
 \sisetup{detect-weight=true}
 \caption{Camvid baseline experiments evaluated on the validation and the test set}
  \centering
 \begin{adjustbox}{width=1\textwidth}
  \begin{tabular}{lSSSSSSSSSS}
    \toprule
    \multirow{2}{*}{Backbone} &
      \multicolumn{2}{c}{$T_{367}$} &
      \multicolumn{2}{c}{$T_{183}$} &
      \multicolumn{2}{c}{$T_{91}$} &
      \multicolumn{2}{c}{$T_{avg}$} \\
      \cmidrule(r){2-3}
      \cmidrule(r){4-5}
      \cmidrule(r){6-7}
      \cmidrule(r){8-9}
       & {Val} & {Test}  & {Val} & {Test}& {Val} & {Test}& {Val} & {Test} &{Param(M)} & {GFLOPs}  \\
      \midrule
     ResNet-18\cite{he2016deep} & 83.3 & 64.9 & 79.7 & \bfseries 63.7 & 70.0 & 56.6 & 77.7 &61.7  & 12.3 & 12.4 \\
     EfficientNet-b1\cite{TanL19}& \bfseries 84.4 & \bfseries 68.4 & 75.0 & 61.3 & 77.0 & 58.8 &  78.8& 62.8 & 7.4 & \bfseries 1.5 \\
     RegNetY-08\cite{RegNet}& 80.4 & 64.3 & 77.7 & 61.4 & 70.9 & 57.8 & 76.3 &61.2 & 7.0 &5.8  \\
     MobileNet-V2\cite{mobileNetV2}& 80.8 & 63.9 & 77.3 & 56.1 & 66.1 & 54.6 & 74.7 &58.2  &4.4 & 4.1 \\
     IkshanaNet-3G & 81.6 & 65.7 & \bfseries 80.0  &62.5  &\bfseries 78.0  & \bfseries 61.2 & \bfseries 79.9 & \bfseries 63.1  &\bfseries 0.5 & 26.0 \\
     \midrule
     ResNet-50\cite{he2016deep}& 78.6 & 61.6 &80.0  &60.3  &\bfseries 78.3  &55.9 &80.0  & 59.3  & 26.7 & 25.0\\
     EfficientNet-b4\cite{TanL19}&82.7  &64.1  &77.7  &62.2  &75.6  & \bfseries 60.5 & 78.7 & 62.3 &18.6 & \bfseries 1.7 \\
     RegNetY-40\cite{RegNet}& 80.8 & 62.0 & 76.4 & 61.0 & 74.9 & 59.2 & 77.4 &60.7 &21.5 & 18.8   \\
     ResNext-50 \cite{Resnext}& 80.1 & 62.6 & 77.3 &56.1  &66.1  &54.6  & 74.5 &57.8 & 26.2 & 25.0  \\
     IkshanaNet-6G & \bfseries 83.3  & \bfseries 67.8 & \bfseries 81.4 &  \bfseries 65.9  &76.0  & 60.0 & \bfseries 80.2 & \bfseries 64.6 & \bfseries 1.8 & 82.0 \\
     \midrule
     ResNet-101\cite{he2016deep}& 81.6 & 63.8 &75.6 & 56.4  & 70.1 & 55.7  & 75.8 &58.6 & 59.3 & 59.9   \\
     EfficientNet-b6\cite{TanL19}& 80.6 &65.0  & 80.3 & 57.8 & \bfseries 77.4 & 60.4 & 79.4 & 61.0 & 42.0 & \bfseries 1.9  \\
     RegNetY-80\cite{RegNet}& 78.5 & 62.0 & 78.2 &63.8  & 66.2 & 53.8 &  74.3&  59.9 & 40.3 & 34.4 \\
     DenseNet-161\cite{huang2017densely}&77.8  & 58.6 & 75.7 & 57.8 & 73.0 &53.8  & 75.5 &56.7 & 43.2 & 43.6    \\
     HRNet-V2\cite{SunZJCXLMWLW19} & 81.1  & 63.6 & 79.1 & 62.9 & 72.9 & 55.0  & 77.7 & 60.5 & 65.9 & 63.5    \\
     U-Net\cite{ronneberger2015u} & 83.0 & 69.5 & 78.0 &62.8  & 76.8  & \bfseries 61.6  &79.3& 64.6 & 31.0 & 130.0  \\
     IkshanaNet-12G & \bfseries 83.9 & \bfseries 70.0 & \bfseries 83.3 & \bfseries  67.1 &  76.5  & 60.6 & \bfseries 81.2  & \bfseries 65.9 & \bfseries 6.5 & 285.0   \\
     \midrule
    IkshanaNet-M & 83.2 &68.5  & 79.9 &62.9  &72.2  & 58.8 & 78.4 & 63.4  & 4.0 & 139.0 \\
    \bottomrule
  \end{tabular}
  \end{adjustbox}
  \label{table-4}
\end{table}
\begin{table}[ht]
 \caption{Camvid multi-scale ablation experiments evaluated on the validation and the test set}
  \sisetup{detect-weight=true}
  \centering
 \begin{adjustbox}{width=1\textwidth}
  \begin{tabular}{lSSSSSSSSSS}
    \toprule
    \multirow{2}{*}{Backbone} &
      \multicolumn{2}{c}{$T_{367}$} &
      \multicolumn{2}{c}{$T_{183}$} &
      \multicolumn{2}{c}{$T_{91}$} &
      \multicolumn{2}{c}{$T_{avg}$} \\
      \cmidrule(r){2-3}
      \cmidrule(r){4-5}
      \cmidrule(r){6-7}
      \cmidrule(r){8-9}
       & {Val} & {Test}  & {Val} & {Test}& {Val} & {Test}& {Val} & {Test} &{Param(M)} & {GFLOPs}    \\
      \midrule
     1S-6Glances & 79.2 & 60.0 & 77.8 & 58.8 & 66.7 & 50.9 & 74.6 & 56.6 & 0.26 & 45.6 \\
     2S-3Glances & 80.1 & 65.6 & 79.5 & 60.1 & 77.2 & 59.5 & 78.9 &61.7 & 0.26 & 23.1\\
     3S-2Glances & \bfseries 82.9 & \bfseries 66.5 & \bfseries 80.9 & \bfseries 62.8 & \bfseries 77.5 & \bfseries 60.8 & \bfseries 80.4 & \bfseries 63.4 & 0.26 & \bfseries 14.0 \\
  \bottomrule
  \end{tabular}
\end{adjustbox}
  \label{table-5}
\end{table}
\subsubsection{Baseline experiments :}
\label{4.3.1}
Here, according to the size of the networks, we classify the total networks into three different sets. \newline
\textbf{Set-$1$} consists of DeeplabV3+ \cite{chen2018encoder} with the encoder networks such as Resnet-18 \cite{he2016deep}, EfficientNet-b1 \cite{TanL19}, RegNetY-08 \cite{RegNet}, MobileNet-V2 \cite{mobileNetV2}, and IkshanaNet-3G (see Section \ref{3.2}). \newline
\textbf{Set-$2$} consists of  DeeplabV3+ \cite{chen2018encoder} with the encoder networks such as Resnet-50 \cite{he2016deep}, EfficientNet-b4 \cite{TanL19}, RegNetY-40 \cite{RegNet}, and ResNext-50 \cite{Resnext}, and IkshanaNet-6G (see Section \ref{3.2}). \newline
\textbf{Set-$3$} consists of DeeplabV3+ \cite{chen2018encoder} with the encoder networks such as Resnet-101 \cite{he2016deep}, EfficientNet-b6 \cite{TanL19}, RegNetY-80 \cite{RegNet},  DeepLabV3 ( DenseNet-161 \cite{huang2017densely}), HRNet-V2 \cite{SunZJCXLMWLW19}, U-Net \cite{ronneberger2015u}, and IkshanaNet-12G (see section \ref{3.2}) \footnote{Same as section \ref{4.2.1}, except for U-Net \cite{ronneberger2015u} and IkshanaNet-12G, we use the ImageNet \cite{deng2009imagenet} pre-trained weights for all the networks in the Set-$3$.}.\newline
Additionally, we include IkshanaNet-main and did not compare it with other networks. By using the same validation, we train each network on three different subsets of the training data, $T_{367}$, $T_{183}$, and $T_{91}$.\newline
In table \ref{table-4}, we provide the mean IoU results evaluated on the validation set, the test set, the average M.IoU score of all the variants, the number of parameters (in Million), and the GFLOPs \cite{GFLOPS}  (calculated the GFLOPs with an input resolution of $1$x$368$x$480$x$3$). From table  \ref{table-4}, we observe the following:\newline
In \textbf{Set-$1$}: (i) IkshanaNet-$3$G outperforms all other networks in the subsets $T_{91}$,  $T_{avg}$, and requires fewer parameters.\newline
(ii) EfficientNet-b1 \cite{TanL19} outperforms other networks in the $T_{367}$ and requires fewer GFLOPs. \newline
In \textbf{Set-$2$}: (i) IkshanaNet-6G outperforms all other networks in the subsets $T_{367}$, $T_{183}$, $T_{avg}$, and requires fewer parameters. \newline
(ii) EfficientNet-b4 \cite{TanL19} outperforms all other networks in the subset $T_{91}$ and requires fewer GFLOPs.\newline
In \textbf{Set-$3$}: (i) IkshanaNet-12G outperforms all other networks in the subsets $T_{367}$, $T_{183}$, $T_{avg}$, and requires fewer parameters. \newline
(ii) U-Net \cite{ronneberger2015u} outperformed other networks in the  subset $T_{91}$ and  EfficientNet-b6 \cite{TanL19} requires fewer GFLOPs than other networks.

\subsubsection{Multi-scale ablation study :}
\label{4.3.2}
Same as Section \ref{4.2.3}, by using the same validation set, we train three different variants of IkshanaNet such as $1$S-$6$G, $2$S-$3$G, and $3$S-$2$G (explained in section \ref{3.2}) on three subsets of the training data ($T_{367}$, $T_{183}$, and $T_{91}$). \newline
In table \ref{table-5}, we provide the mean IoU results evaluated on the validation set, the test set, the average score of all variants, the parameters, and the GFLOPs. We calculate the GFLOPs with an input resolution of $1$x$368$x$480$x$3$.\newline
From table \ref{table-5}, we observe that, the IkshanaNet-3S-2G network outperforms all other networks in all the subsets ($T_{367}$,  $T_{183}$,  $T_{91}$, $T_{avg}$), and requires fewer GFLOPs. The results are similar to the section \ref{4.2.3} (Table \ref{Table-3}), demonstrating the importance of multi-scale information.

\section{Validity threats}
\label{validity-threats}
\noindent
(i) Most of the existing works \cite{chen2018encoder,zhao2017pyramid,10.1007/978-3-030-58539-6_11} used a mini-batch size of $8$ and SyncBN \cite{rotabulo2017place,Zhang_2018_CVPR} for training. However, due to the limited availability of the computing resources, we train all the networks with a mini-batch size of $2$. Due to this reason, we cannot directly compare the performance of our method with the state-of-the-art methods.\newline
(ii) In this work, even though the training data splits are reproducible, the performance of the networks trained on subsets of the training data might depend upon the fact that ``how well the subset represents the whole dataset?". If we use a different random seed to generate the splits, then the exact behavior may or may not be expected.\newline
(iii) In this work, through multi-scale ablation experiments, we observe that multi-scale information is often necessary to improve the performance of the networks. By observing the images in Cityscapes, and CamVid datasets,  it is evident that the images consist of multi-scale objects. However, this phenomenon might not be valid to other datasets, where there exist no multi-scale objects.

\section{Conclusion}
\label{6}
In this work, we attempt to bridge the gap between the current vision DNNs and the human visual system by proposing a novel hypothesis of human scene understanding and a  neural-inspired CNN architecture that learns representations at full-scale resolution.\newline
The empirical results illustrate the effectiveness of our method on entire and few data samples compared to the baselines. Also, through multi-scale ablation studies, we observe that using multi-scale information improves the performance of IkshanaNet by reducing the computational complexity.\newline
Moreover, we observe that our method is just an improvement over the baselines, and it is still dependent on the data. Hence, it is nowhere close to the human visual system. Therefore, a better-performing and computationally efficient architectures based on the Ikshana hypothesis will be studied in the future work.\newline
Furthermore, we hope that our hypothesis inspires future generation of neural inspired vision architectures. 

\bibliographystyle{unsrt} 
\bibliography{main.bib}


\newpage

\appendix

\section{Appendix}
\subsection{Cityscapes experiments}
\subsubsection{Baseline experiments}
\begin{figure}[ht]
\centering     
\subfigure{\includegraphics[width=65mm]{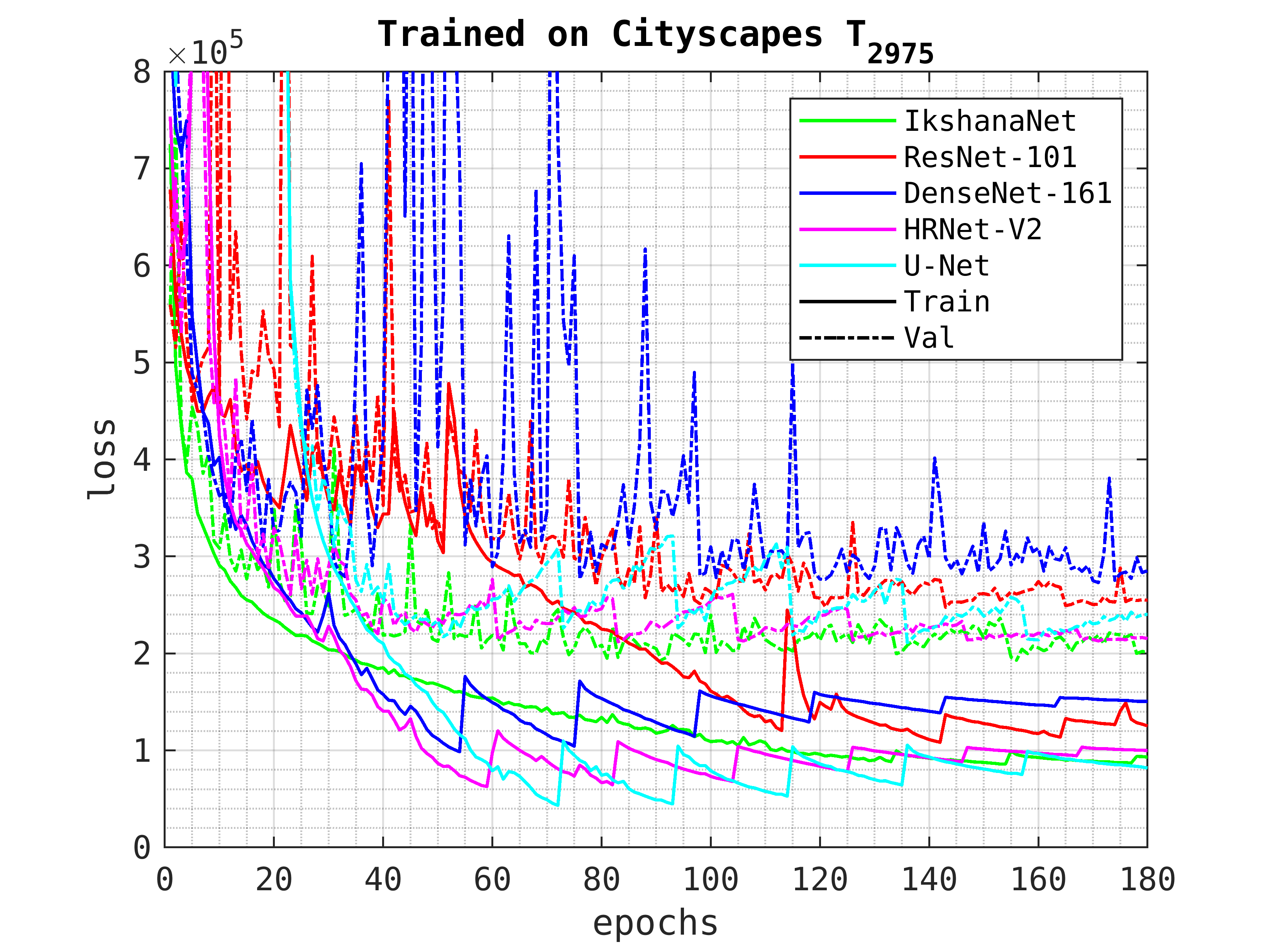}}
\subfigure{\includegraphics[width=65mm]{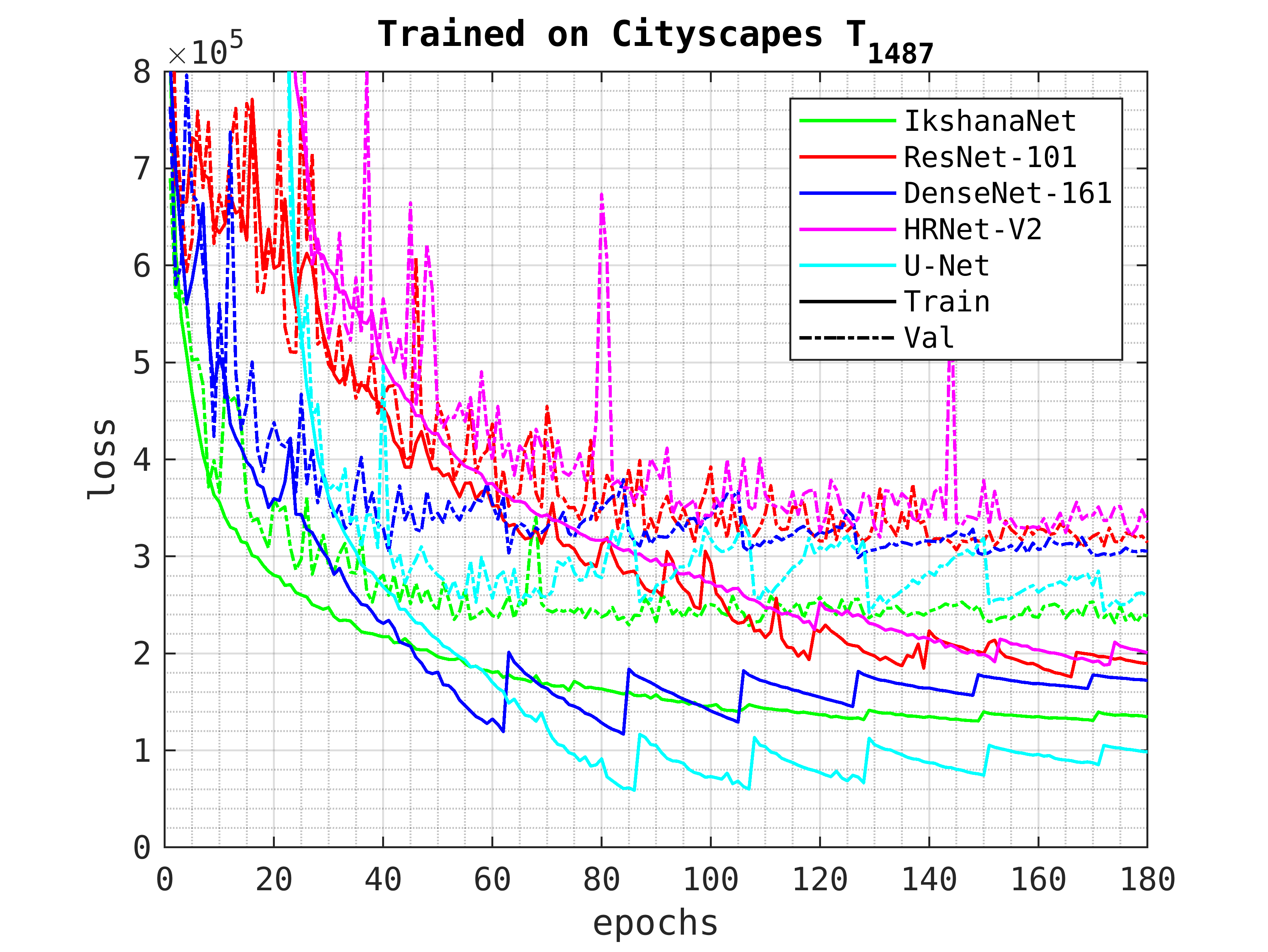}}
\subfigure{\includegraphics[width=65mm]{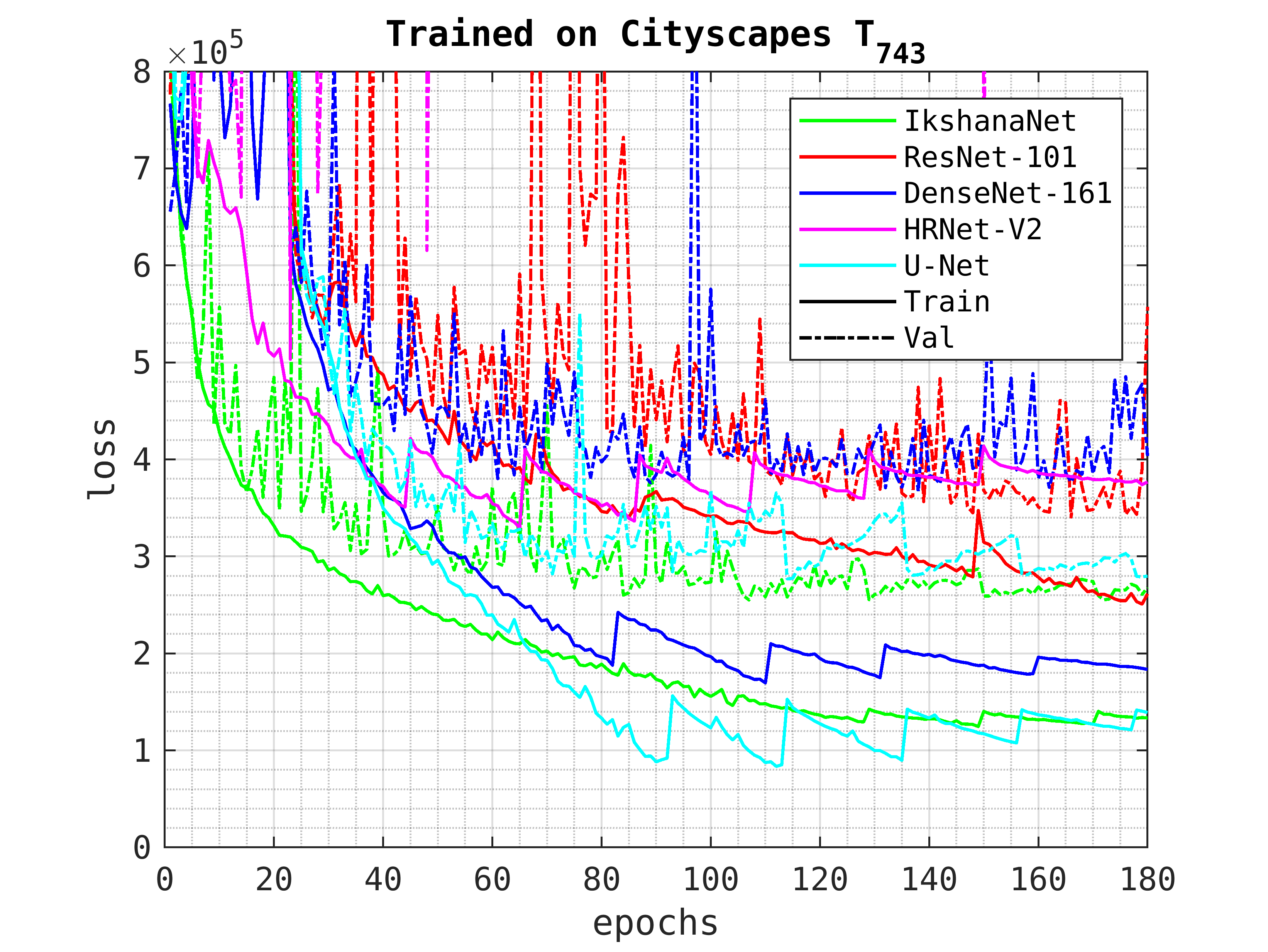}}
\subfigure{\includegraphics[width=65mm]{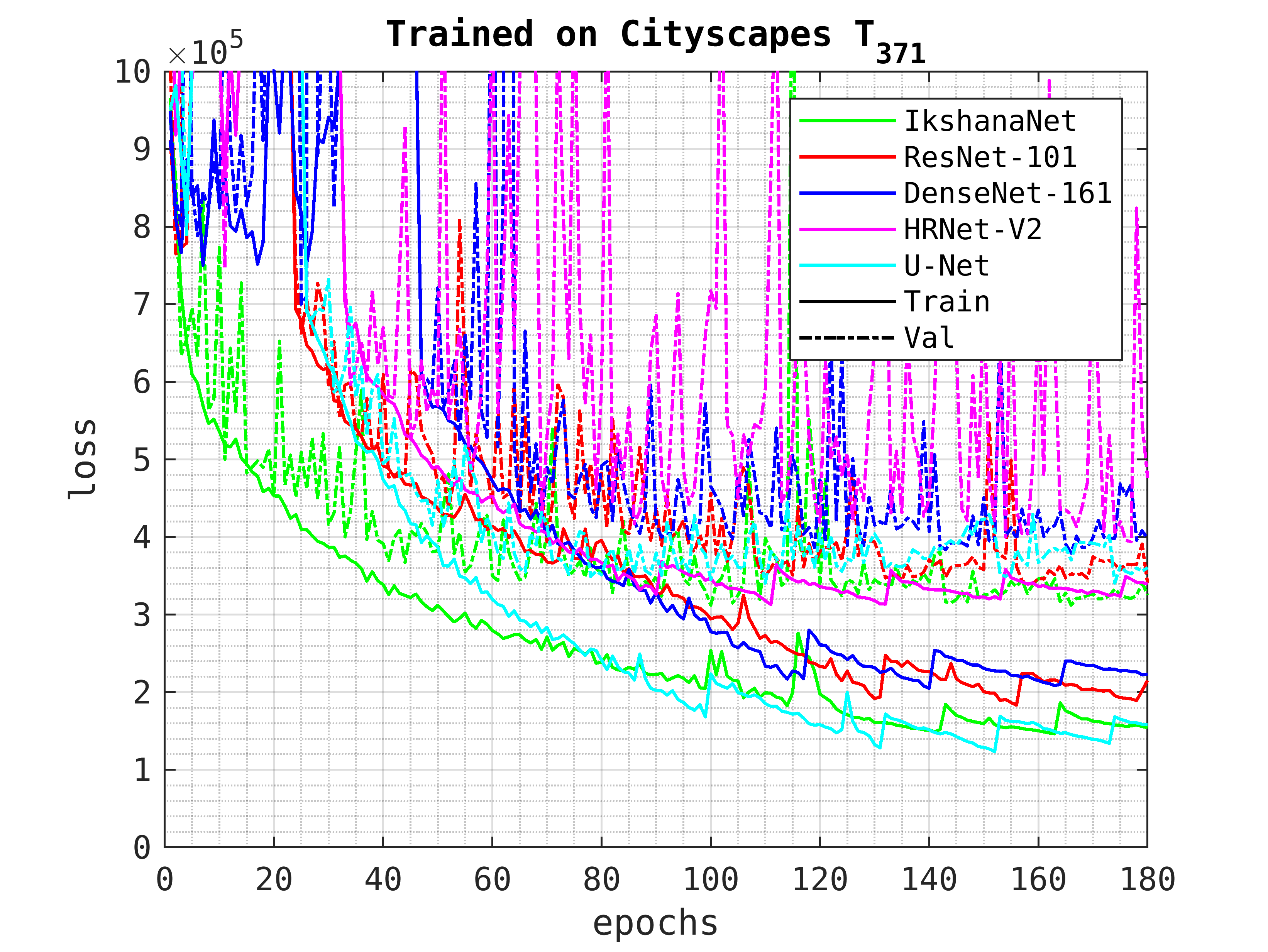}}
\subfigure{\includegraphics[width=65mm]{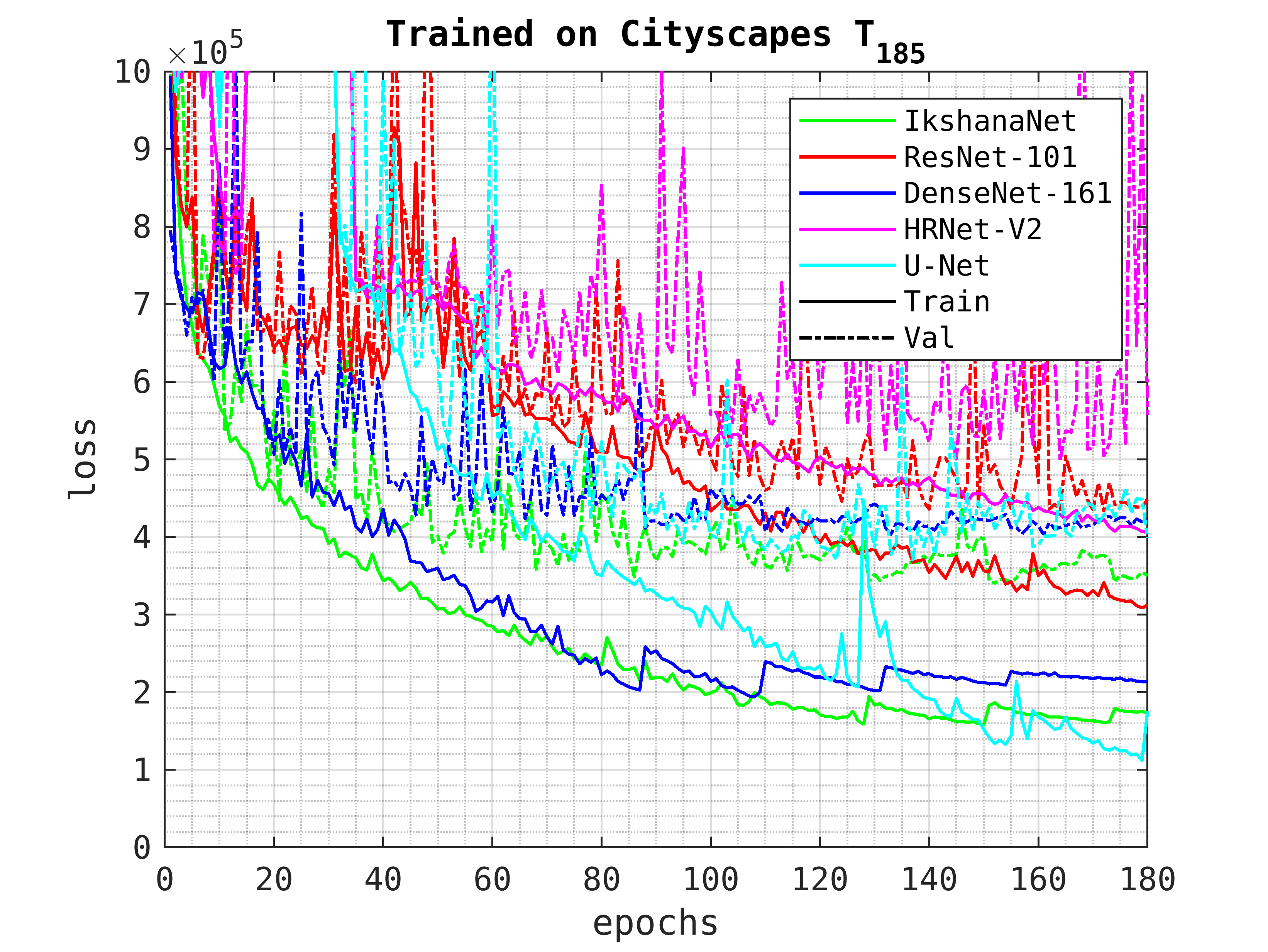}}
\subfigure{\includegraphics[width=65mm]{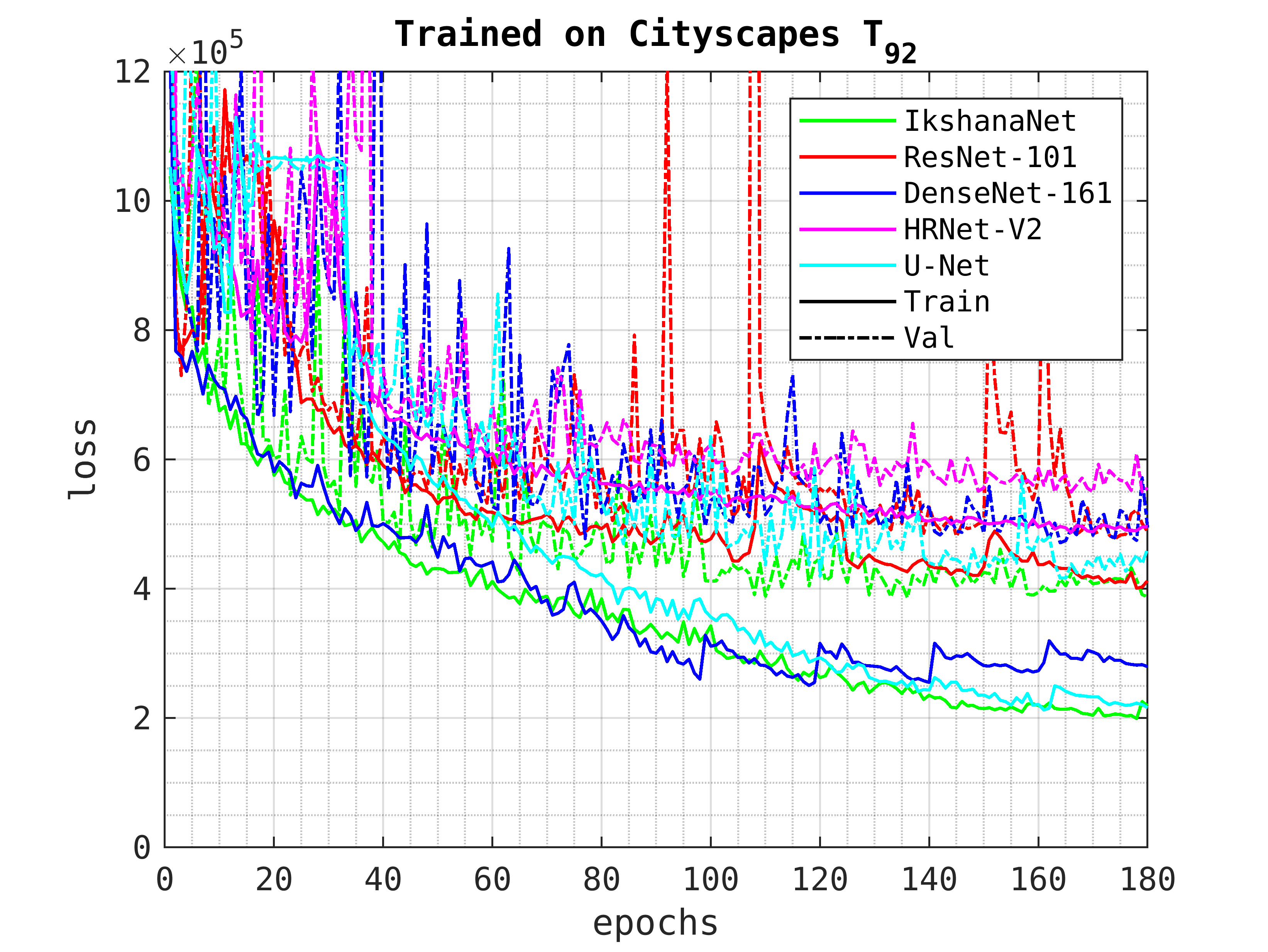}}
\caption{Cityscapes Baseline experiments training plots}
\label{C1}
\end{figure}

For ease of visualization, we separate the training plots of ResNet-101, DenseNet-161,  HRNet-V2, and U-Net from the data ablation study and combine them with baseline experiments. In Figure \ref{C1}, we present the training plots of the networks trained on six subsets of training data ($T_{2975}$, $T_{1487}$, $T_{743}$, $T_{371}$, $T_{185}$, and $T_{92}$).

\subsubsection{Data ablation study}
\begin{figure}[ht]
\centering     
\subfigure{\includegraphics[width=65mm]{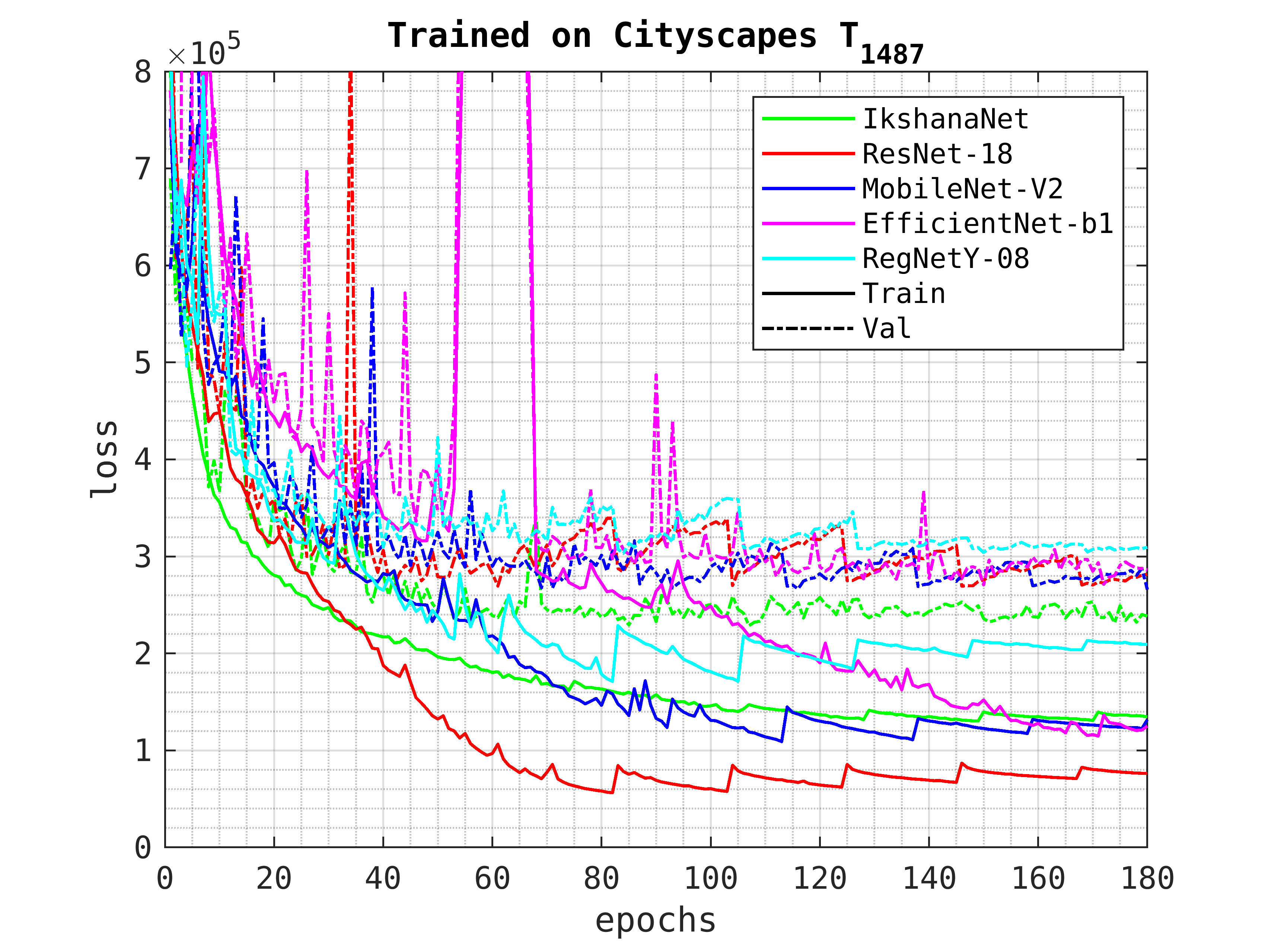}}
\subfigure{\includegraphics[width=65mm]{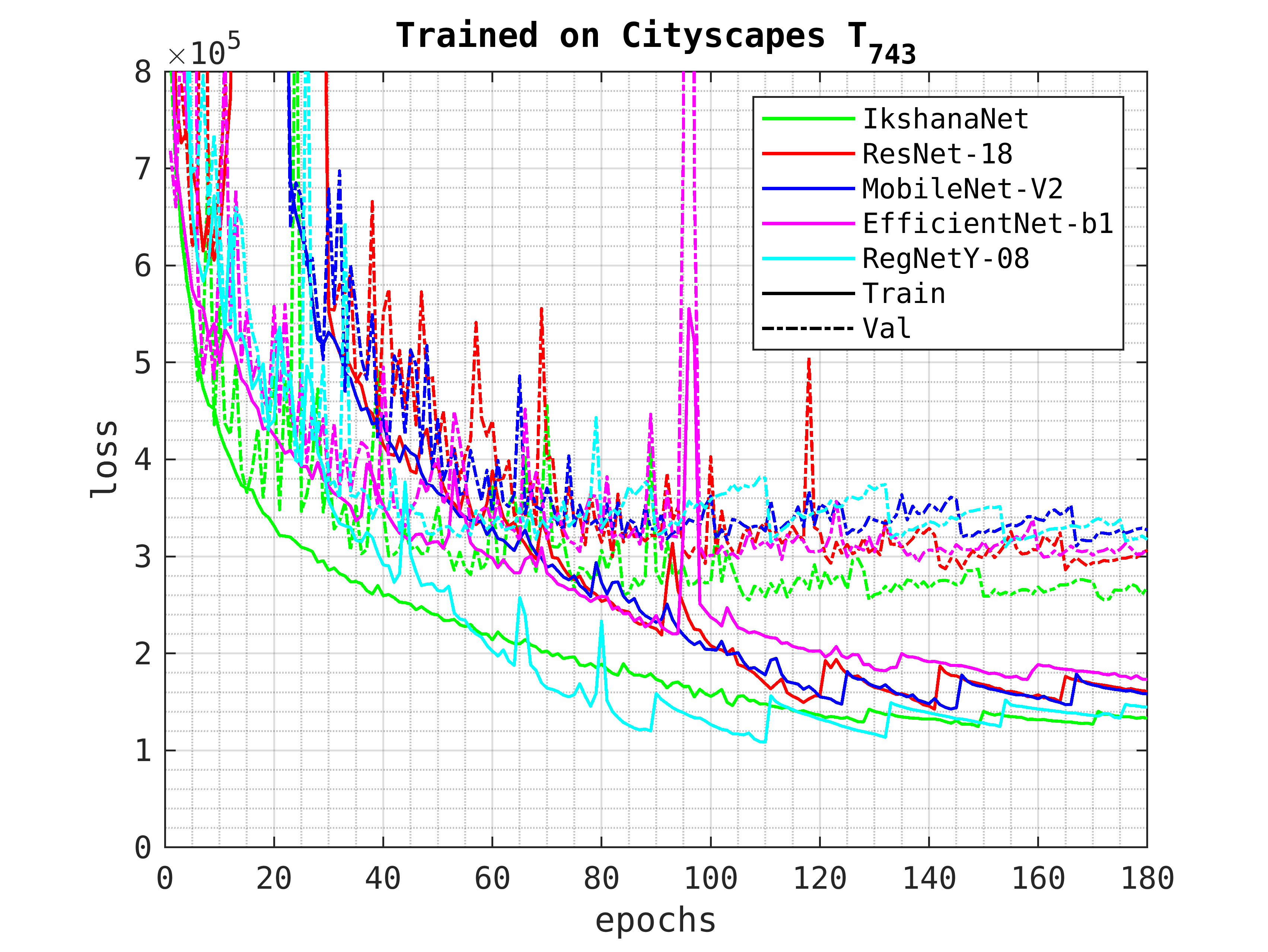}}
\subfigure{\includegraphics[width=65mm]{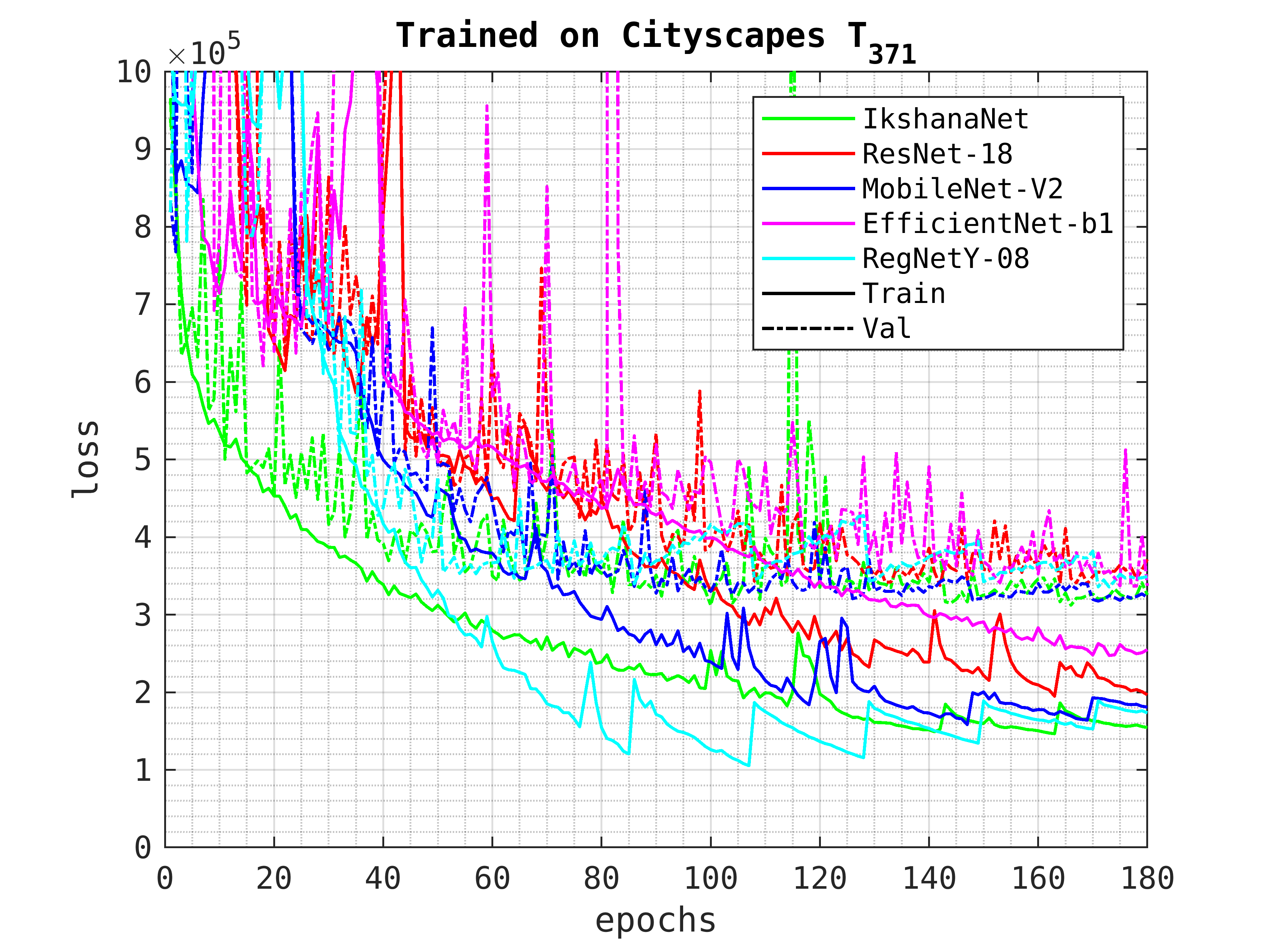}}
\subfigure{\includegraphics[width=65mm]{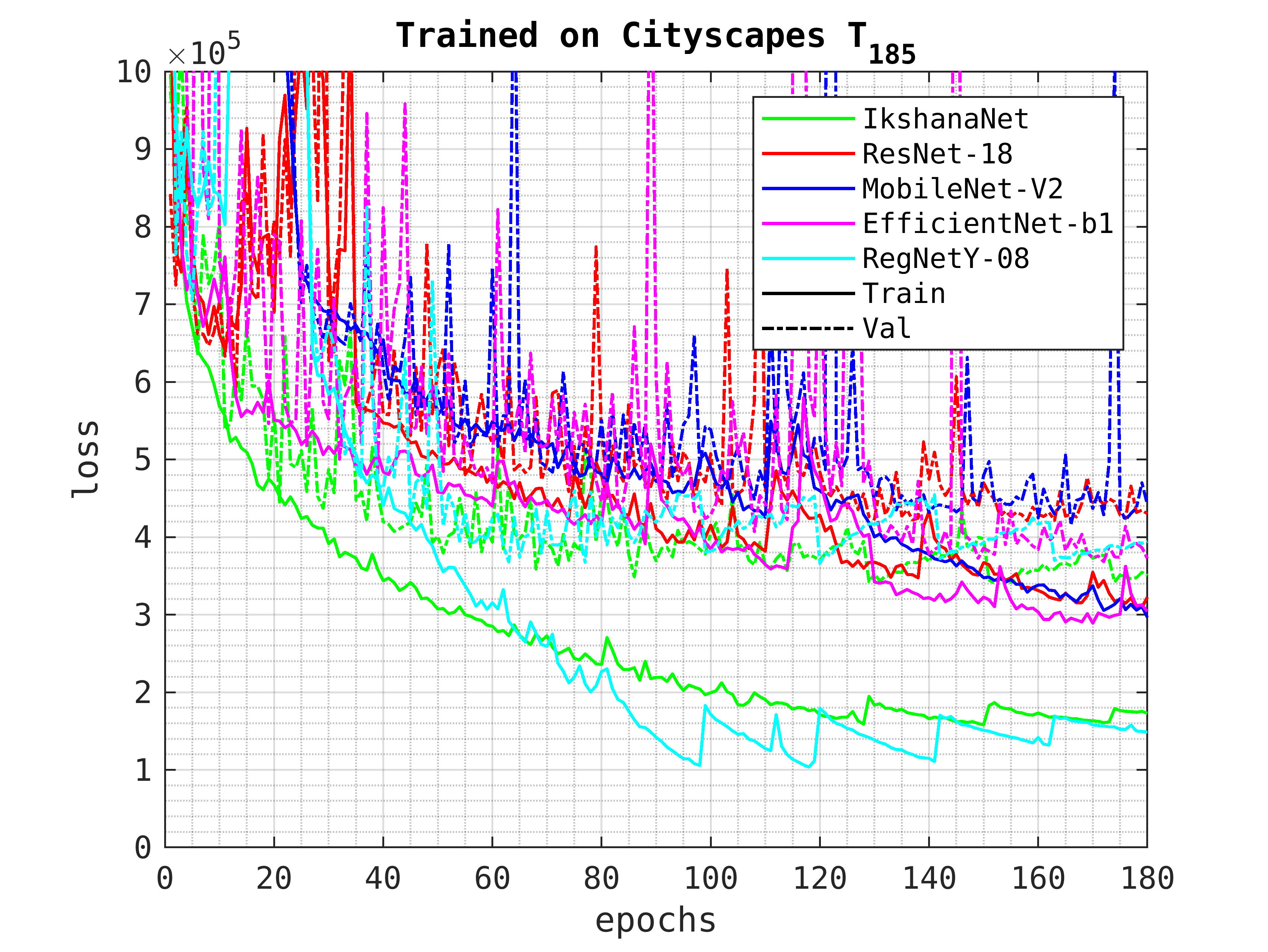}}
\subfigure{\includegraphics[width=65mm]{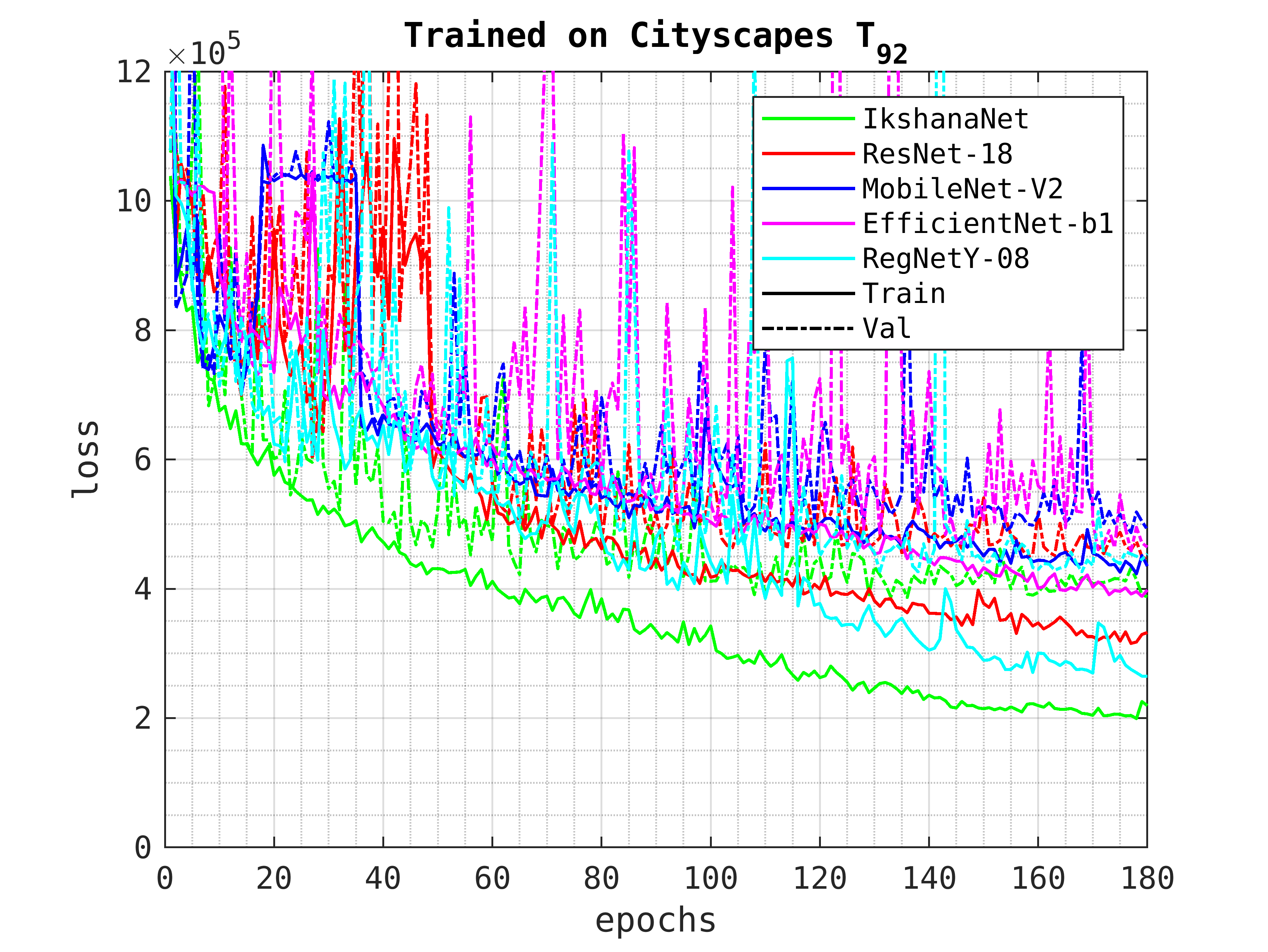}}
\caption{Cityscapes data ablation study training plots}
\label{C2}
\end{figure}

In Figure \ref{C2}, we present the training plots of the networks Resnet-18, MobileNet-V2, EfficientNet-b1 , RegNetY-08 , and IkshanaNet-main,  trained on five subsets of training data ($T_{1487}$, $T_{743}$, $T_{371}$, $T_{185}$, and $T_{92}$).
In table \ref{a-table-2}, we present the class-wise IoU results of these networks.  

\subsubsection{Muti-scale ablation study}
\begin{figure}[ht]
\centering     
\subfigure{\includegraphics[width=65mm]{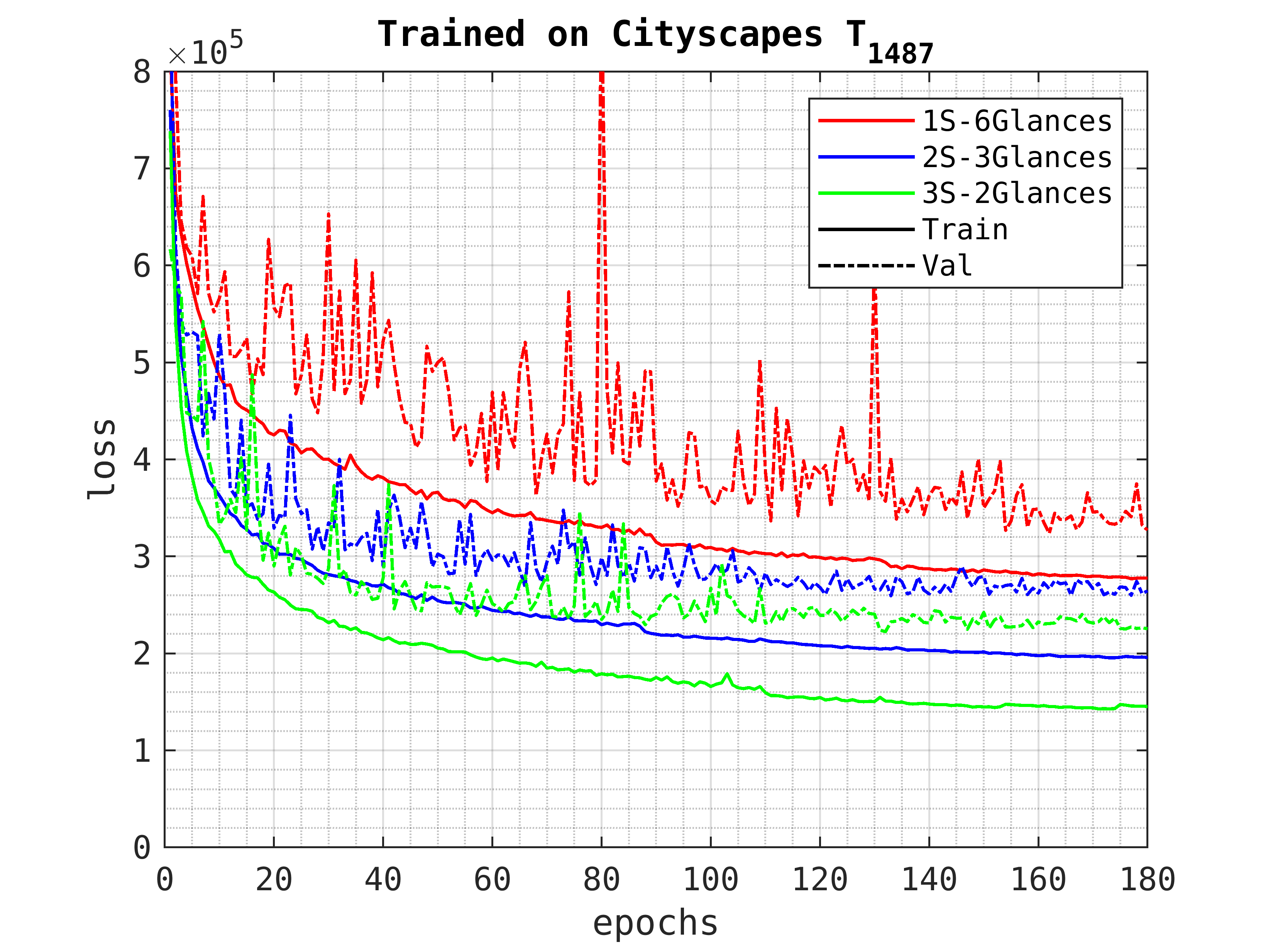}}
\subfigure{\includegraphics[width=65mm]{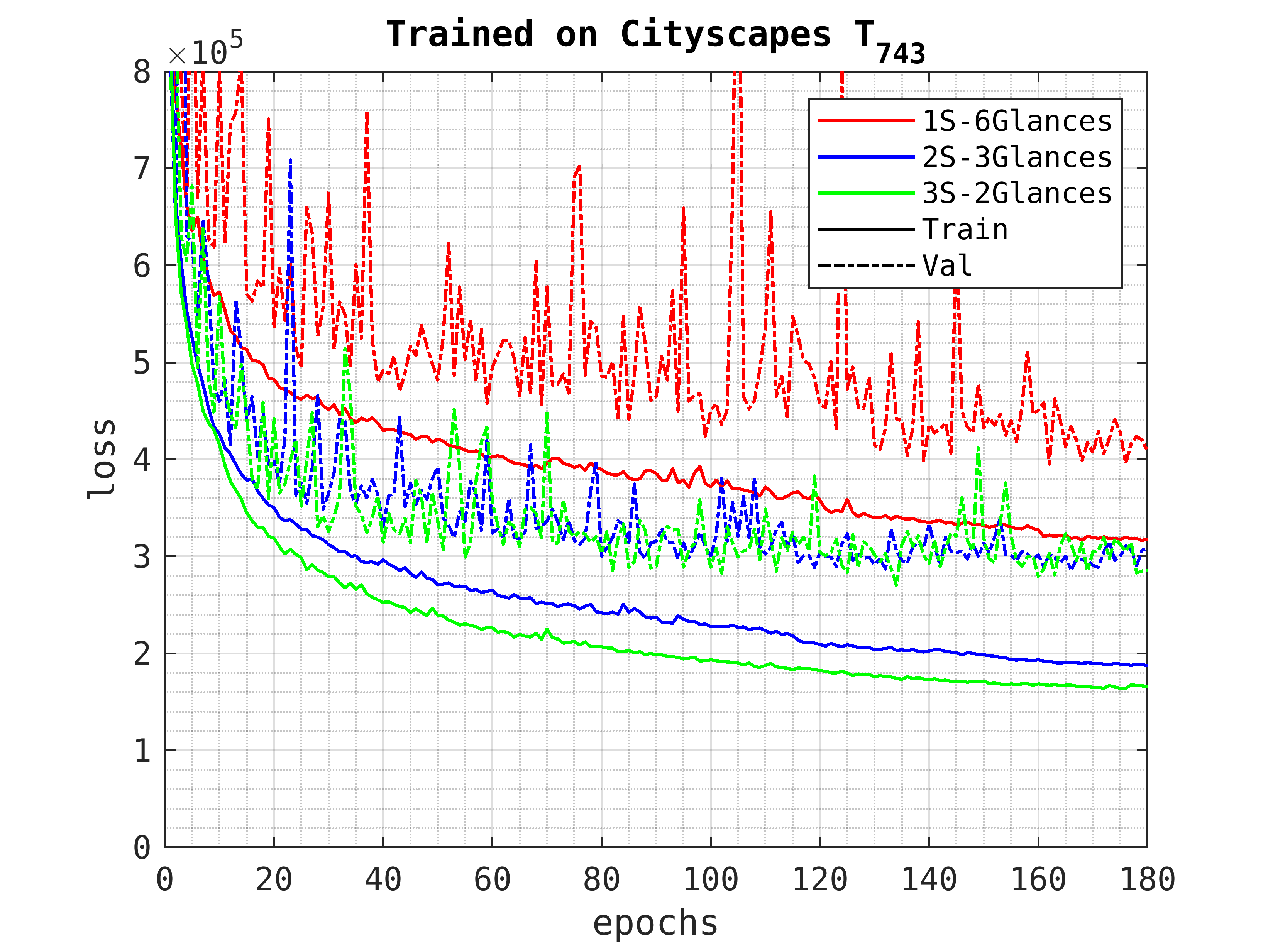}}
\subfigure{\includegraphics[width=65mm]{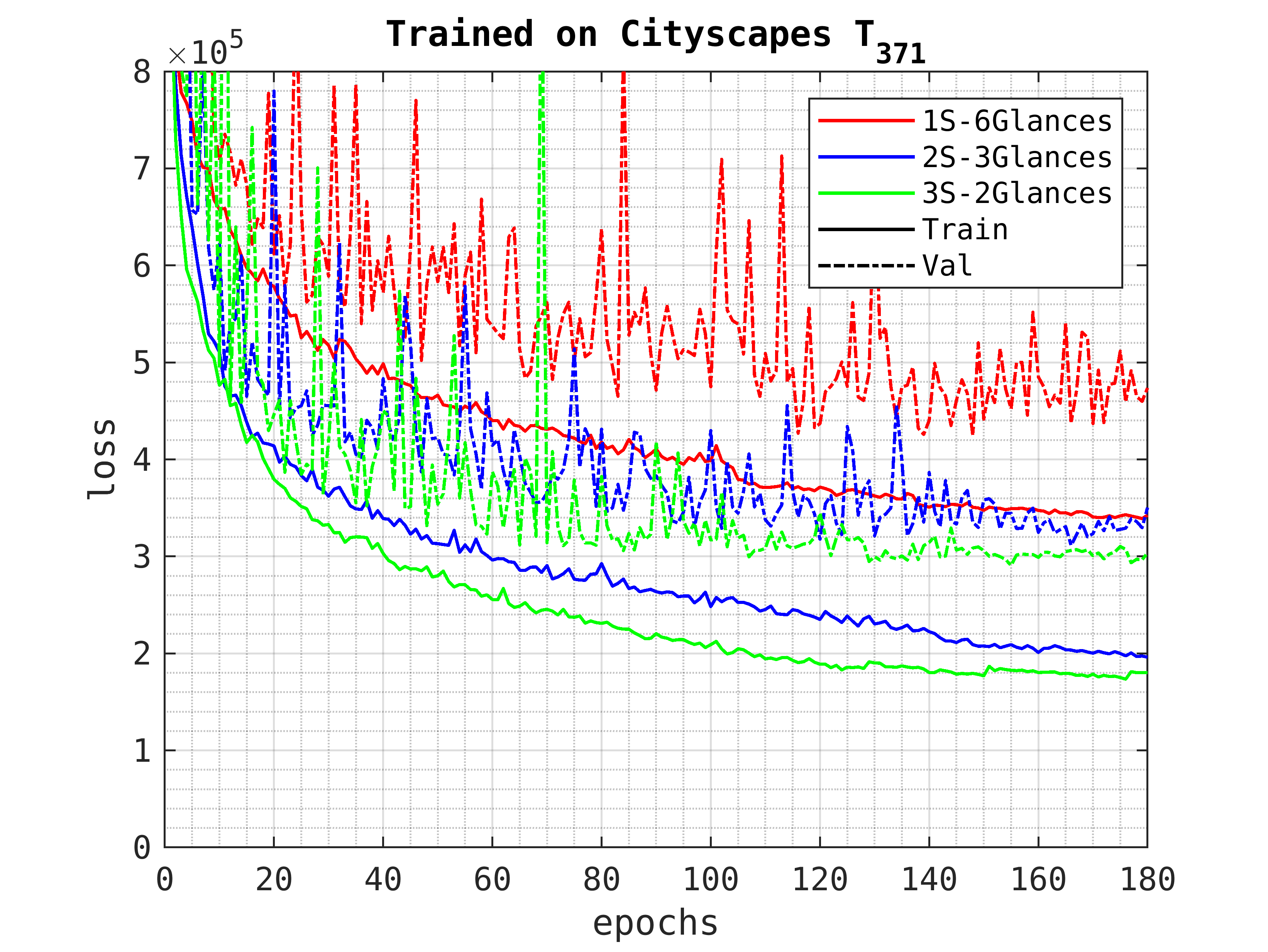}}
\subfigure{\includegraphics[width=65mm]{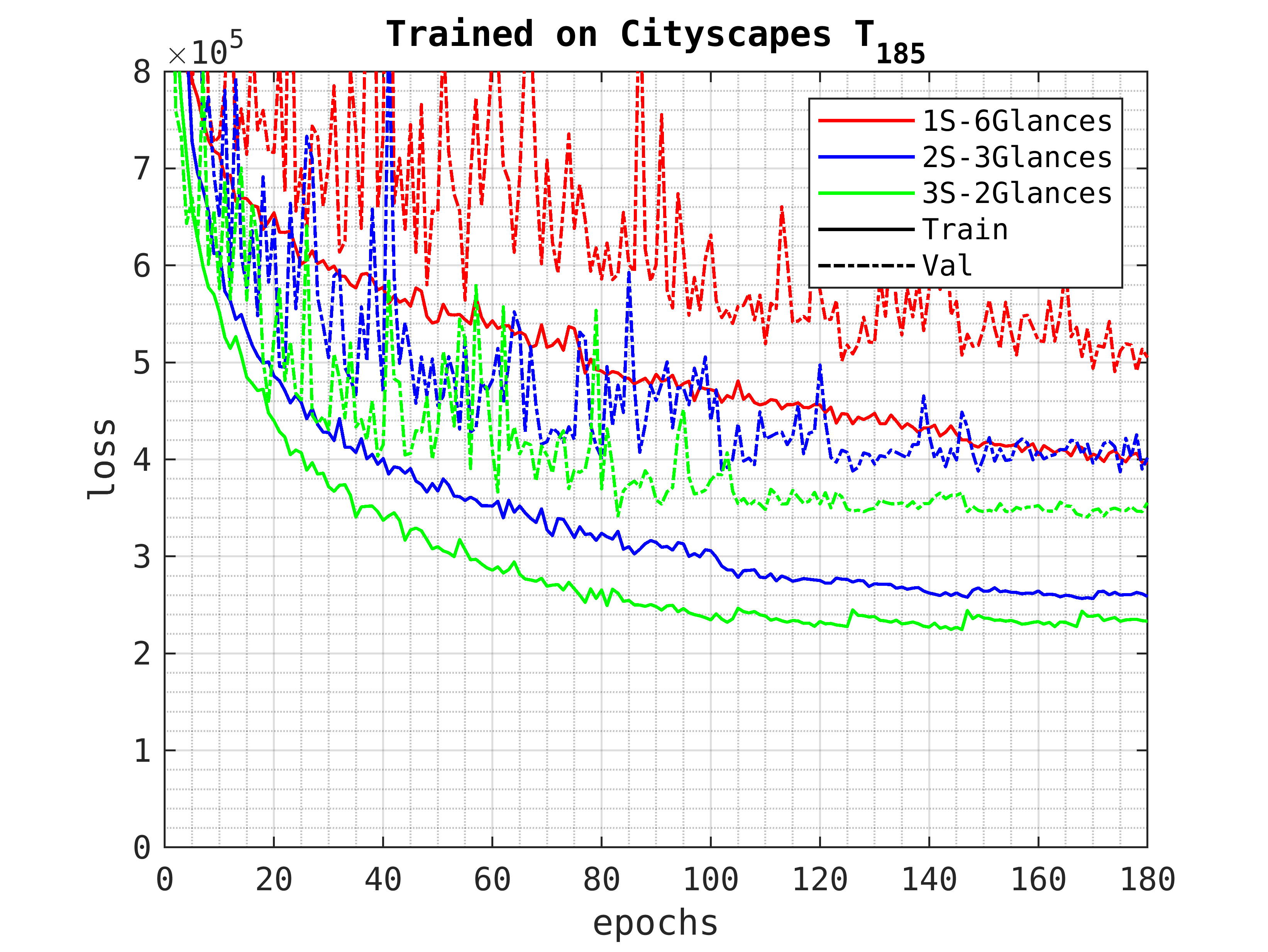}}
\subfigure{\includegraphics[width=65mm]{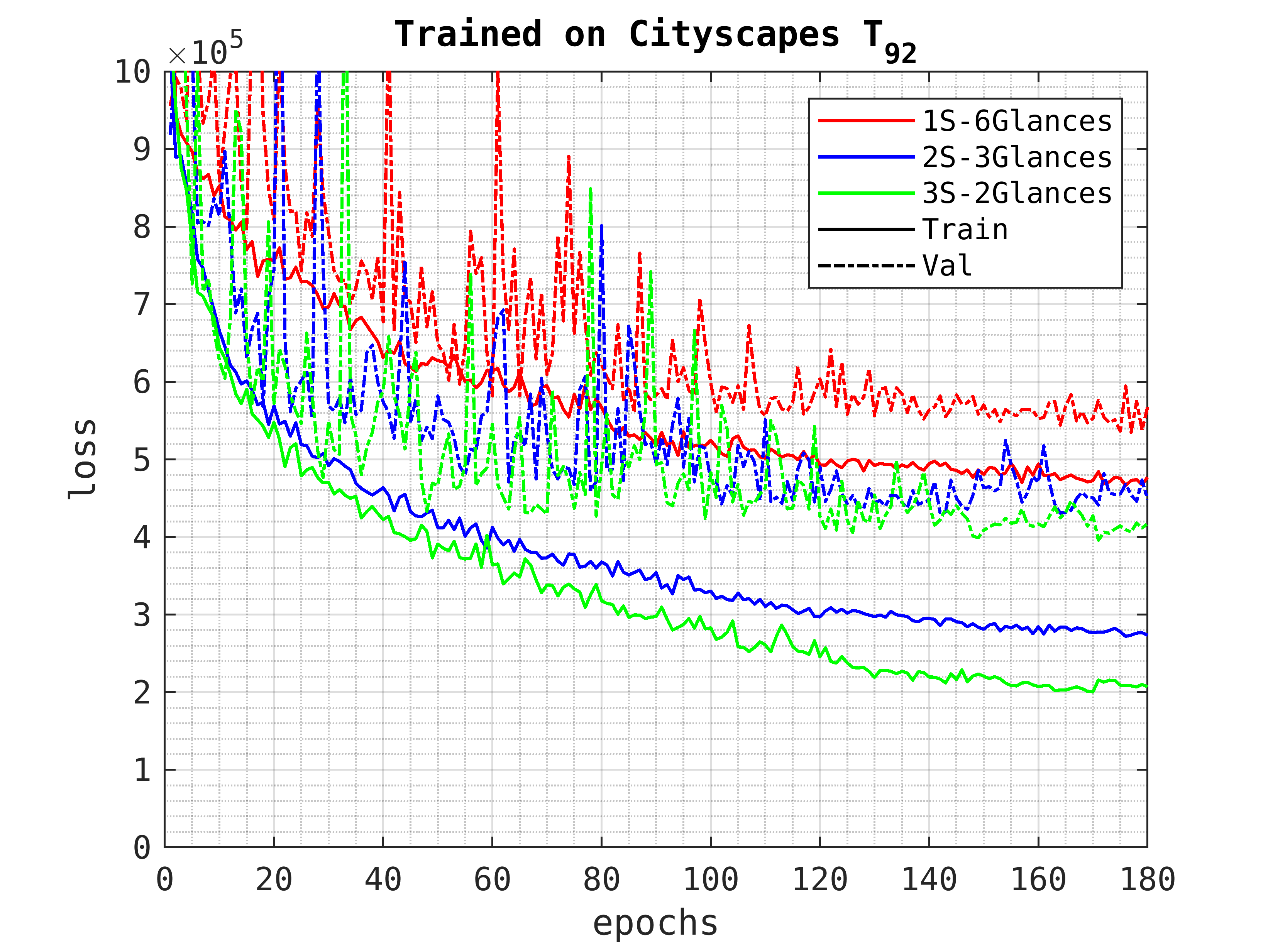}}
\caption{Cityscapes multi-scale ablation study training plots}
\label{C3}
\end{figure}

In Figure \ref{C3}, we present the training plots of the networks IkshanaNet-1S-6G,  IkshanaNet-2S-3G, and IkshanaNet-3S-2G trained on five subsets of training data ($T_{1487}$, $T_{743}$, $T_{371}$, $T_{185}$, and $T_{92}$).\newline
In table \ref{a-table-3}, we present the class-wise IoU results of these networks.

\subsection{CamVid experiments}
\subsubsection{Baseline experiments}
\begin{figure}[ht]
\centering     
\subfigure{\includegraphics[width=65mm]{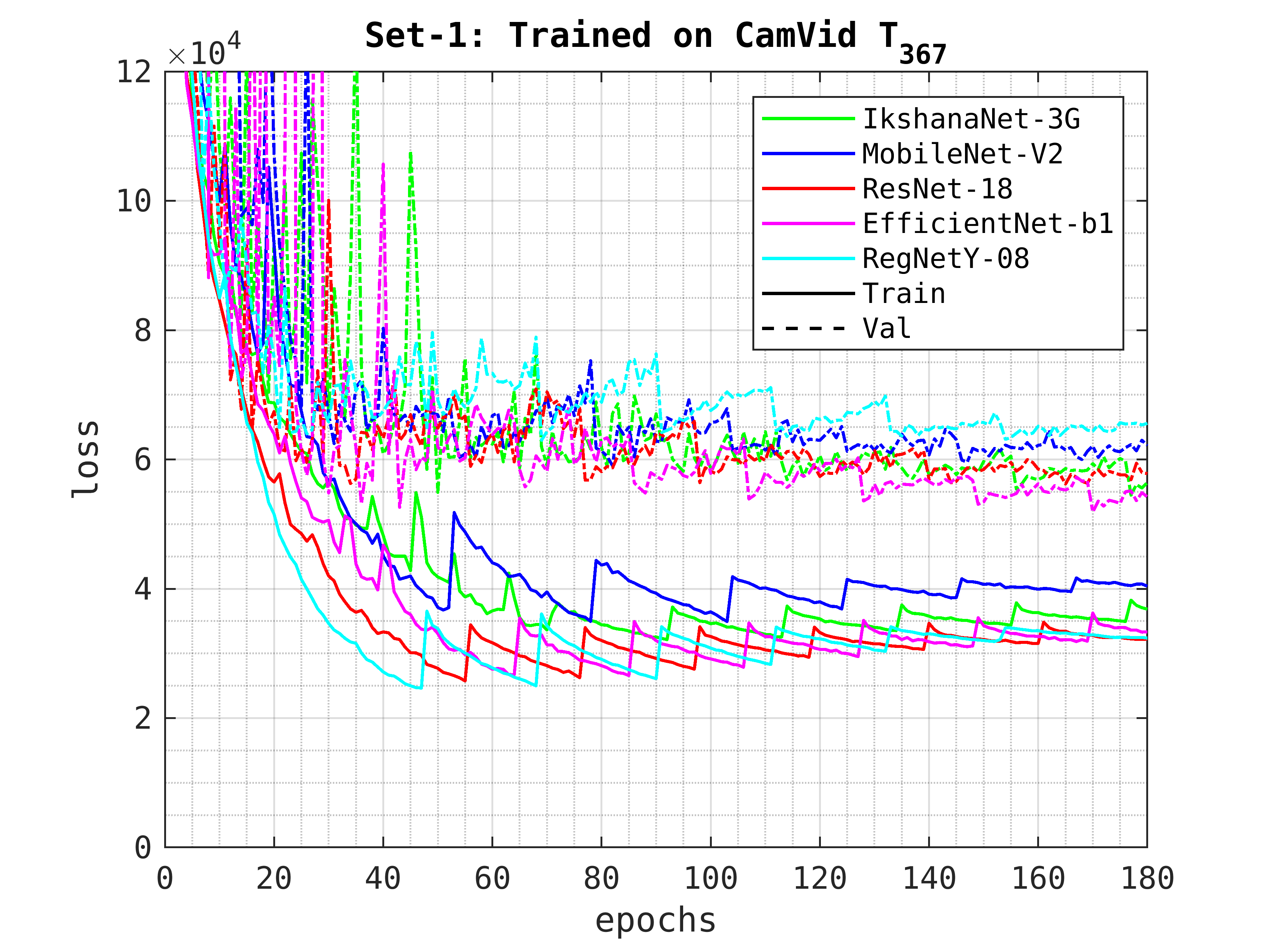}}
\subfigure{\includegraphics[width=65mm]{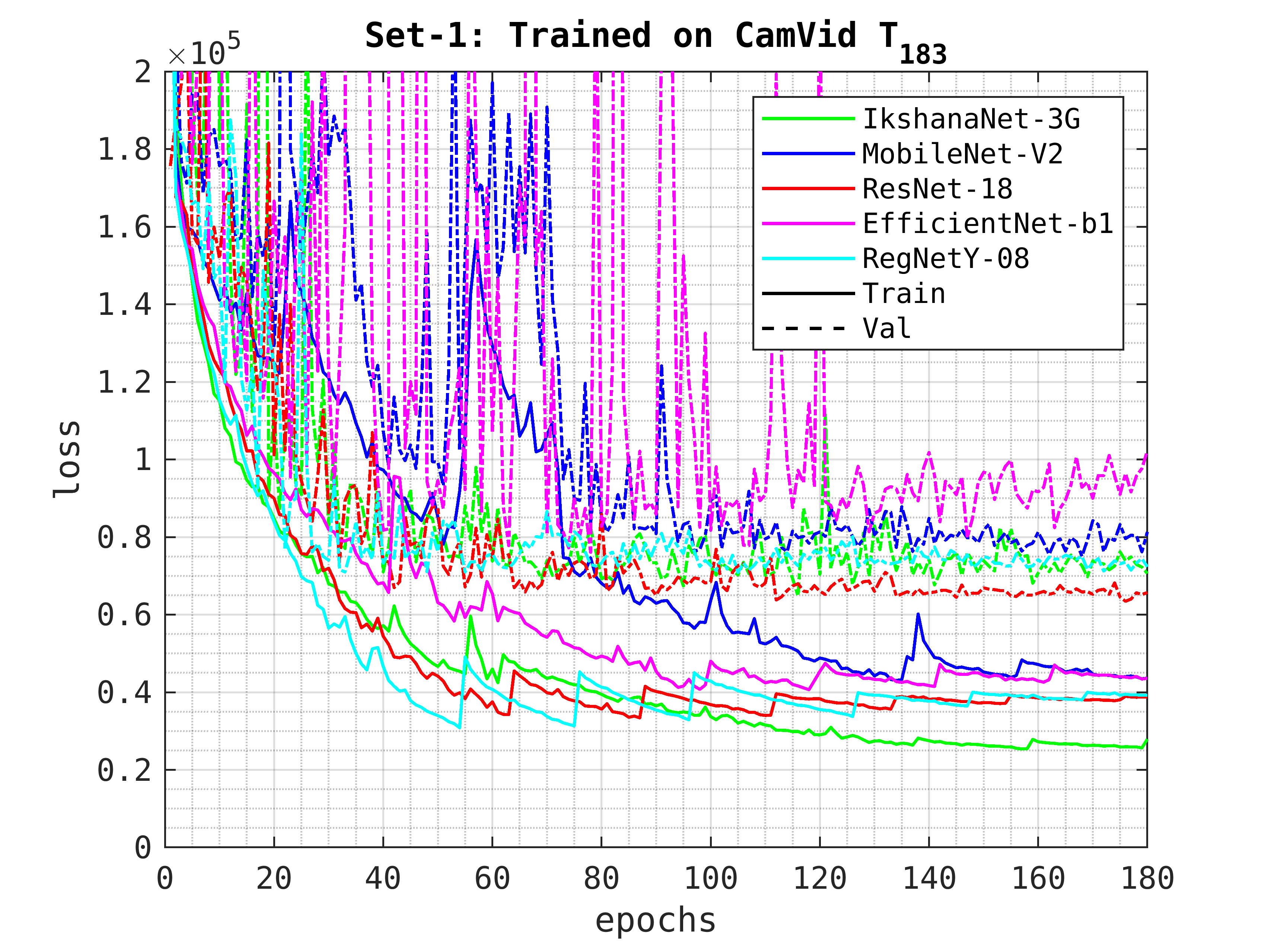}}
\subfigure{\includegraphics[width=65mm]{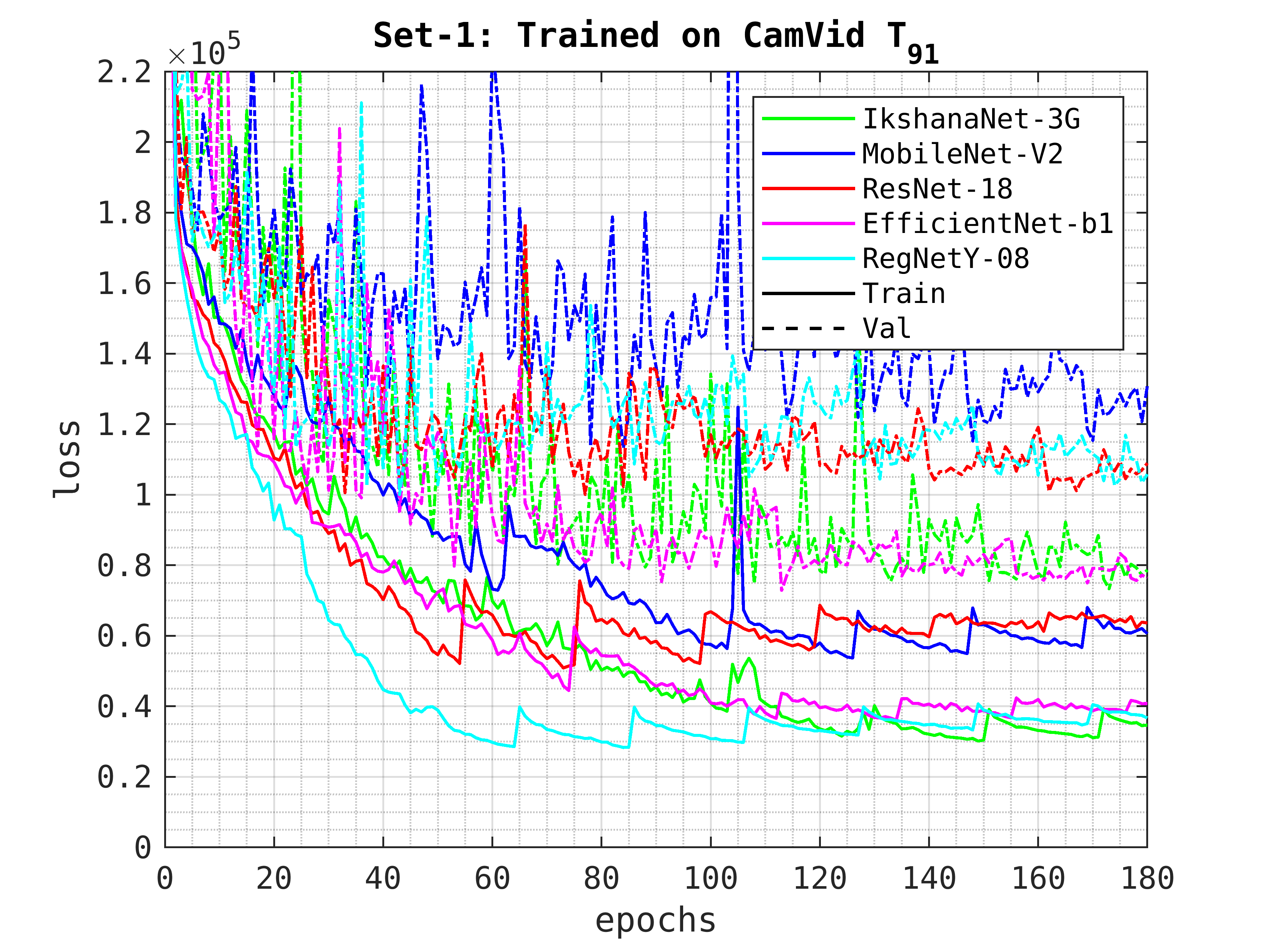}}
\caption{CamVid Set-1 Baseline experiments training plots}
\label{C4}
\end{figure}

\begin{figure}[ht]
\centering
\subfigure{\includegraphics[width=65mm]{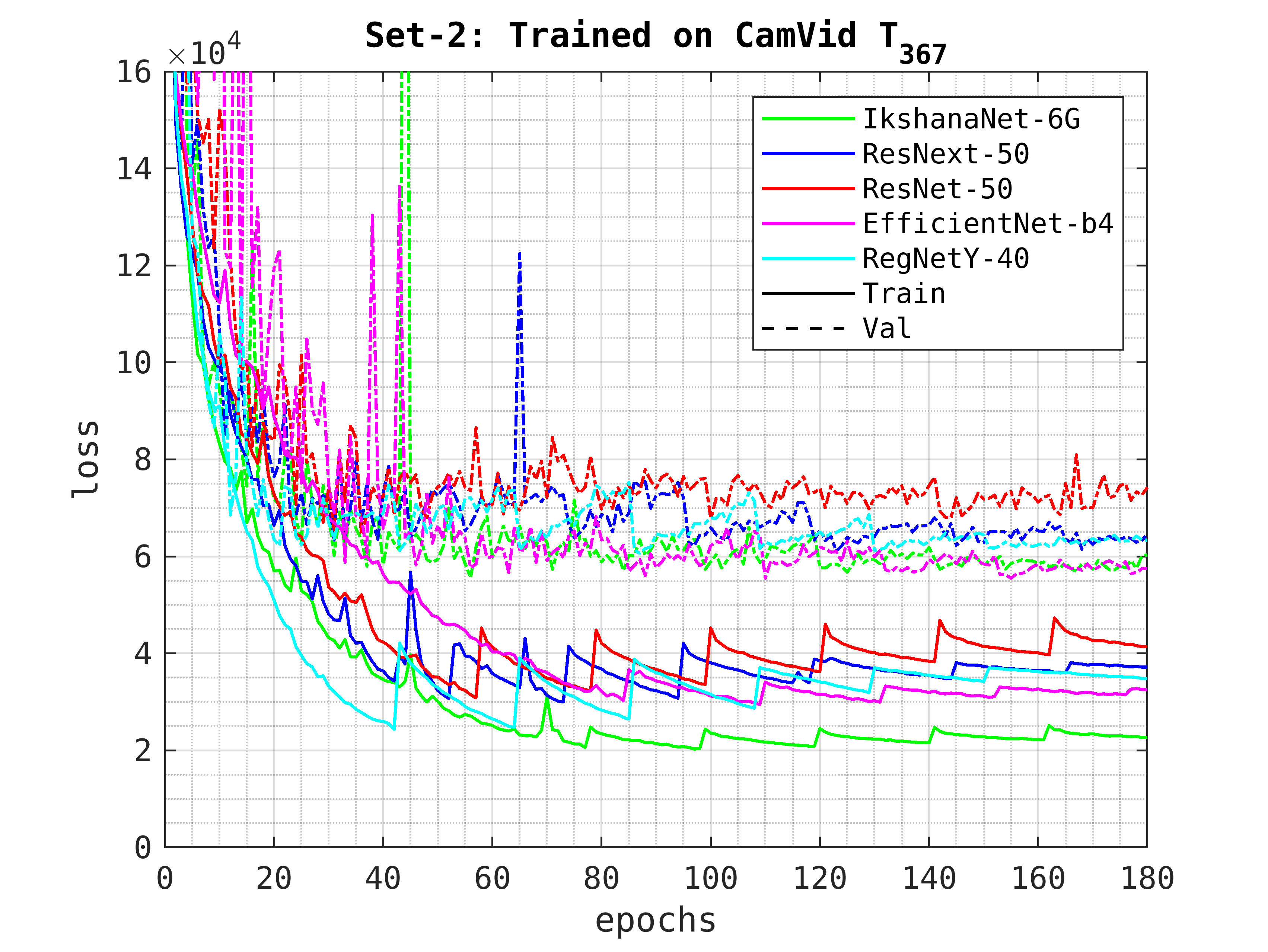}}
\subfigure{\includegraphics[width=65mm]{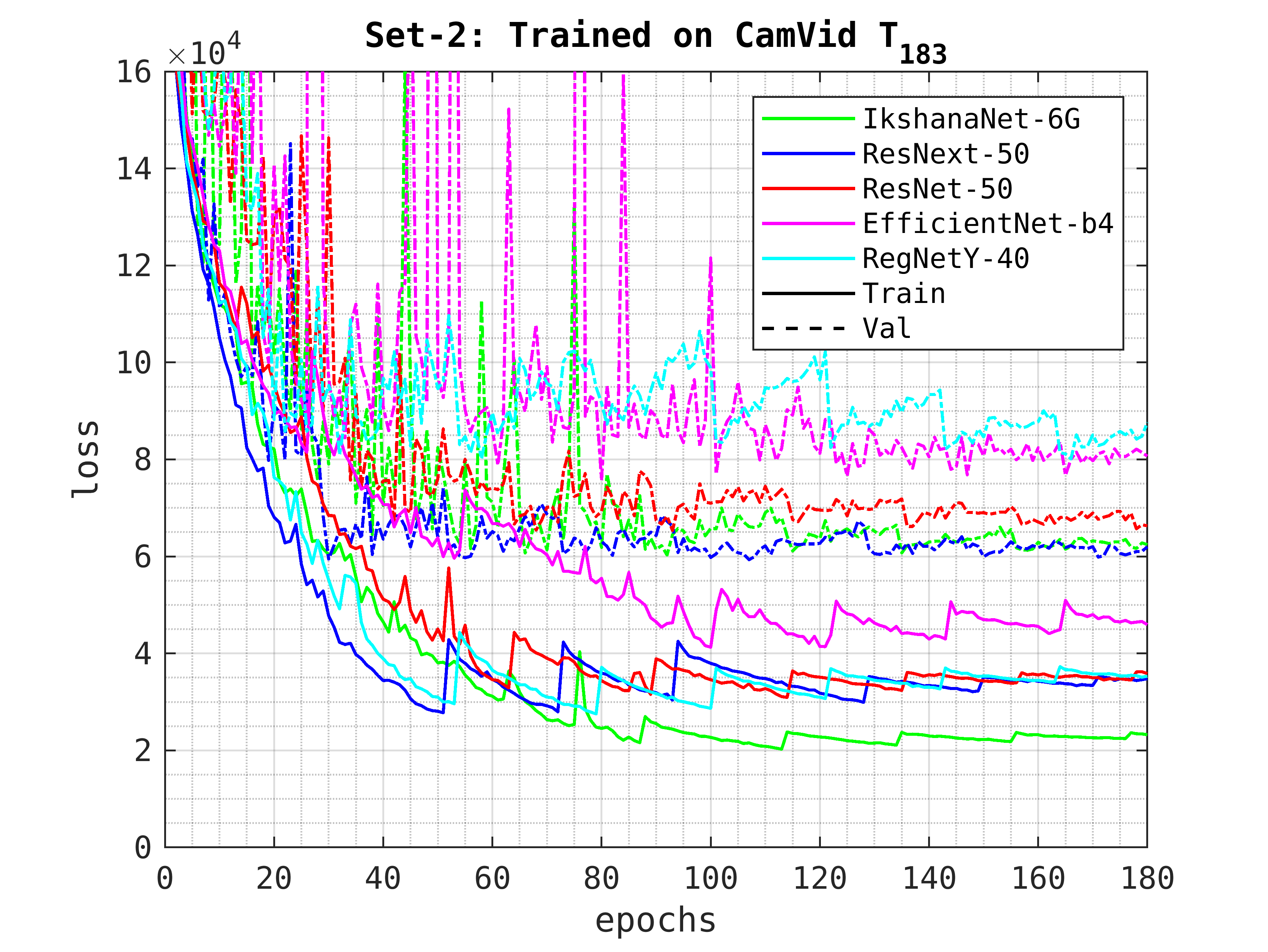}}
\subfigure{\includegraphics[width=65mm]{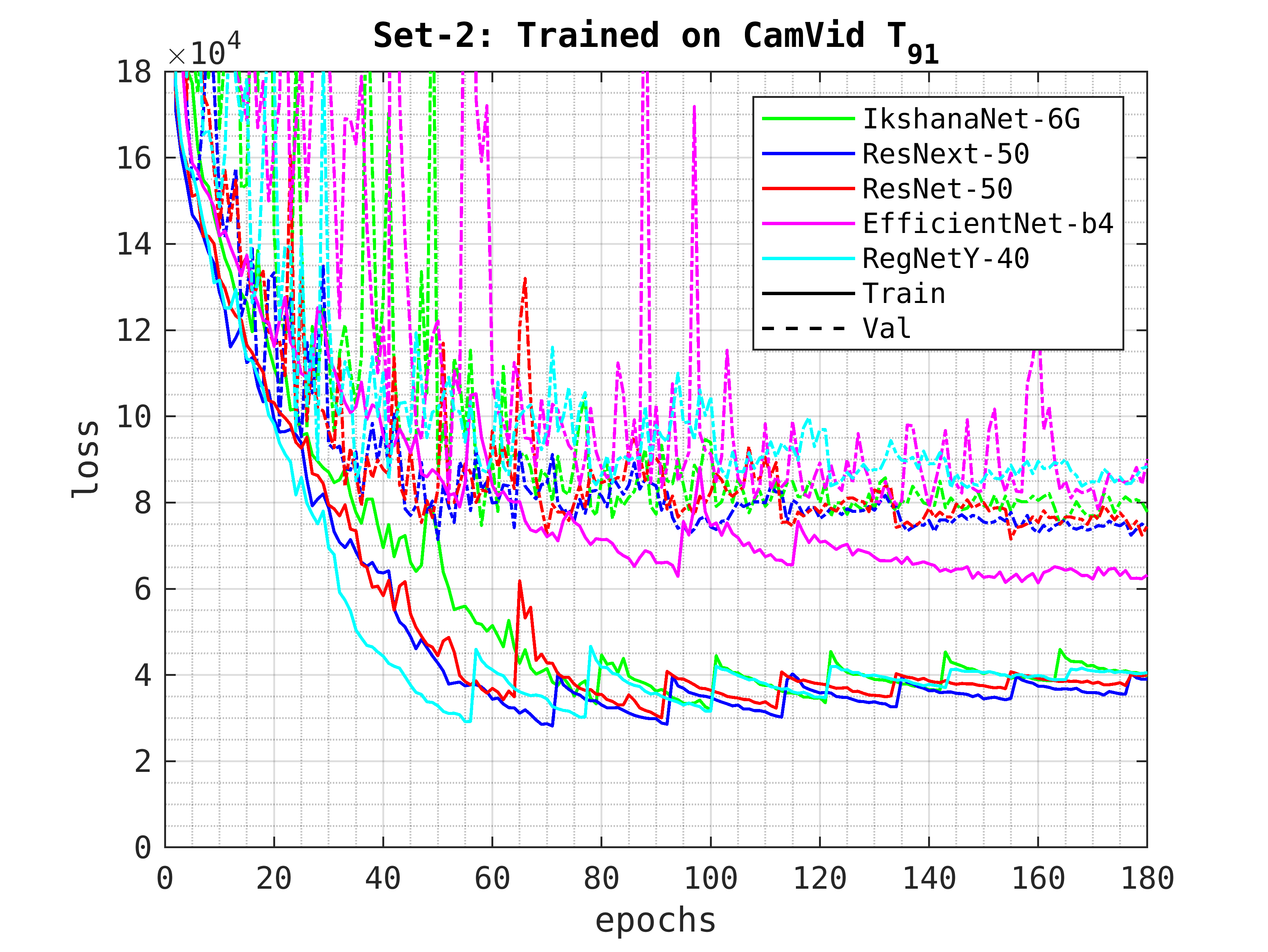}}    
\caption{CamVid Set-2 Baseline experiments training plots}
\label{C5}
\end{figure}

\begin{figure}[ht]
\centering
\subfigure{\includegraphics[width=65mm]{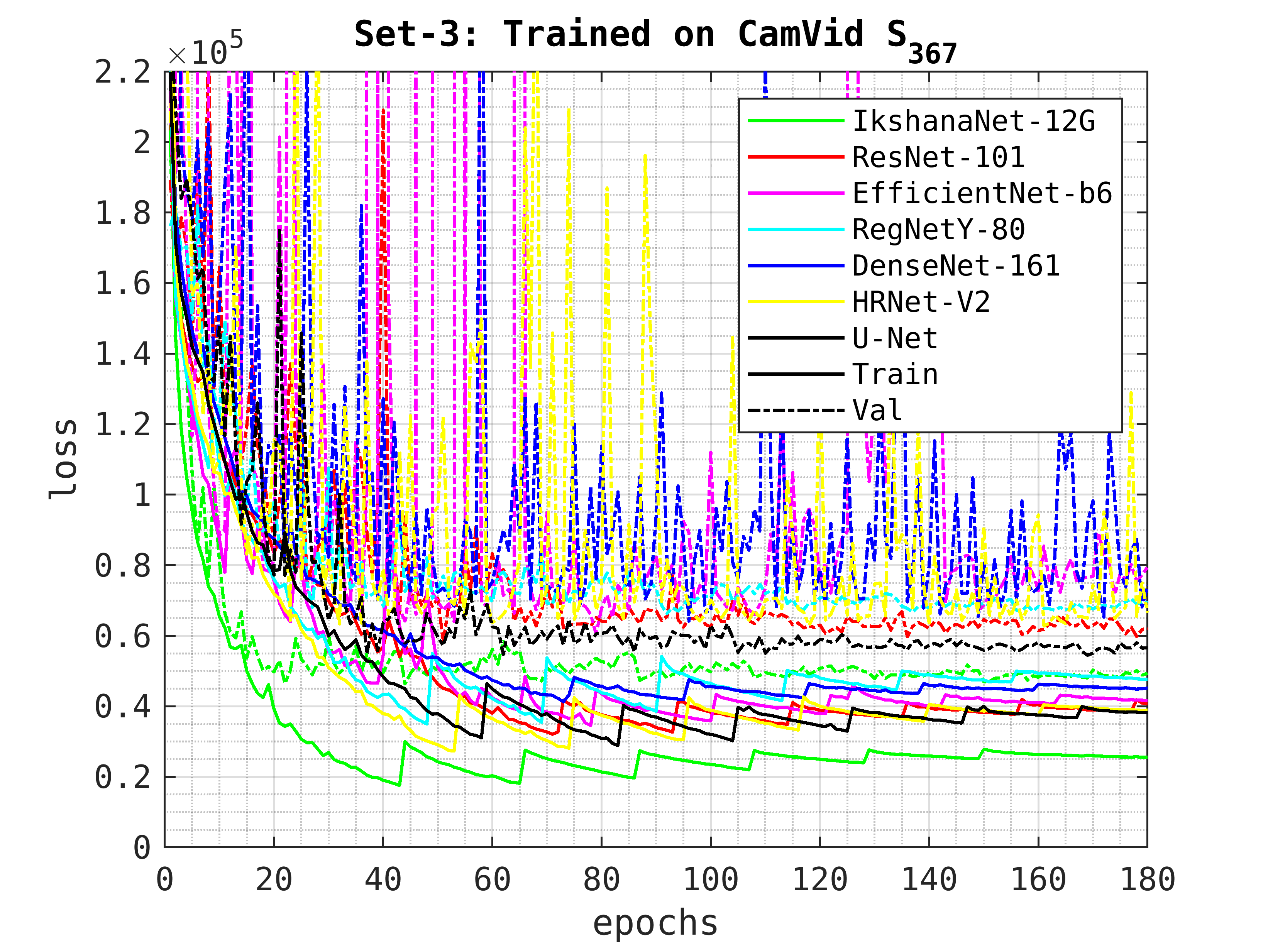}}
\subfigure{\includegraphics[width=65mm]{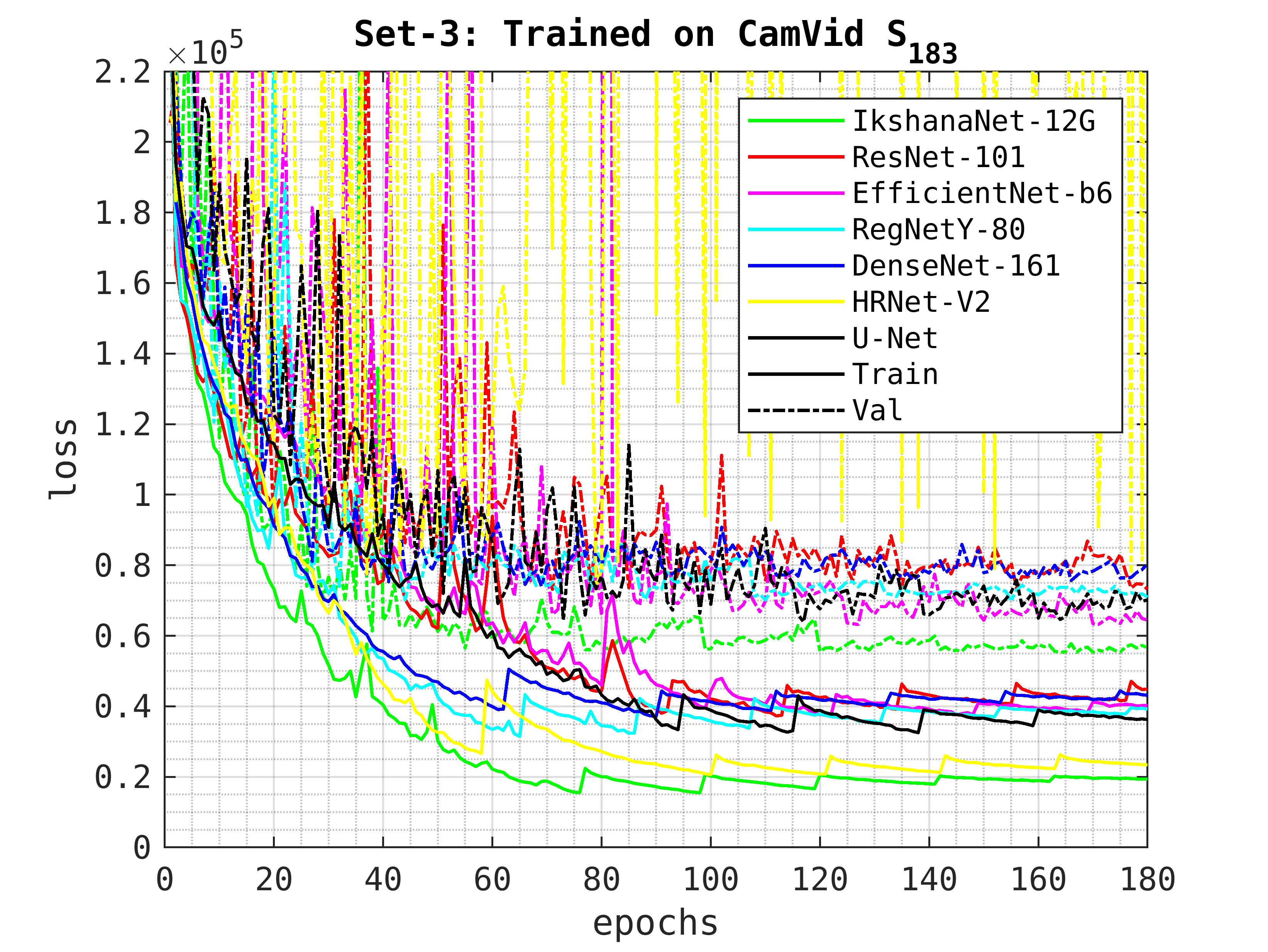}}
\subfigure{\includegraphics[width=65mm]{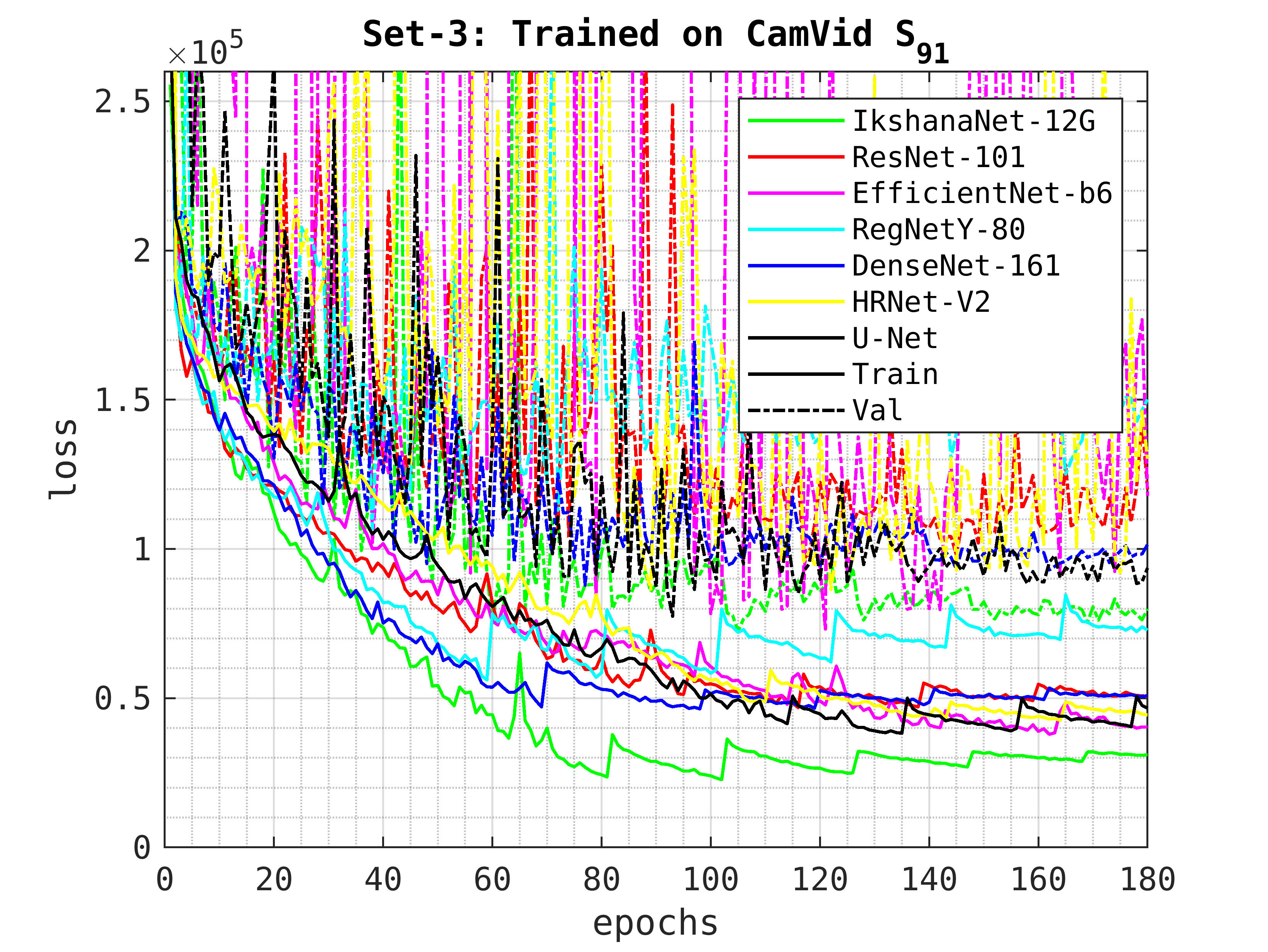}}  
\caption{CamVid Set-3 Baseline experiments training plots}
\label{C6}
\end{figure}

In Figure \ref{C4}, we present the training plots of the networks ResNet-18, EfficientNet-b1, RegNetY-08, MobileNet-V2, and IkshanaNet-3G, trained on three subsets of training data ($T_{367}$, $T_{183}$, and $T_{91}$).\newline
In Figure \ref{C5}, we present the training plots of the networks ResNet-50, EfficientNet-b4, RegNetY-40, ResNext-50, and IkshanaNet-6G, trained on three subsets of training data ($T_{367}$, $T_{183}$, and $T_{91}$).\newline
In Figure \ref{C6}, we present the training plots of the networks ResNet-101, EfficientNet-b6, RegNetY-80, DenseNet-161, HRNet-V2, U-Net, and IkshanaNet-12G, trained on three subsets of training data ($T_{367}$, $T_{183}$, and $T_{91}$).\newline

\subsubsection{Muti-scale ablation study}
\begin{figure}[ht]
\centering     
\subfigure{\includegraphics[width=65mm]{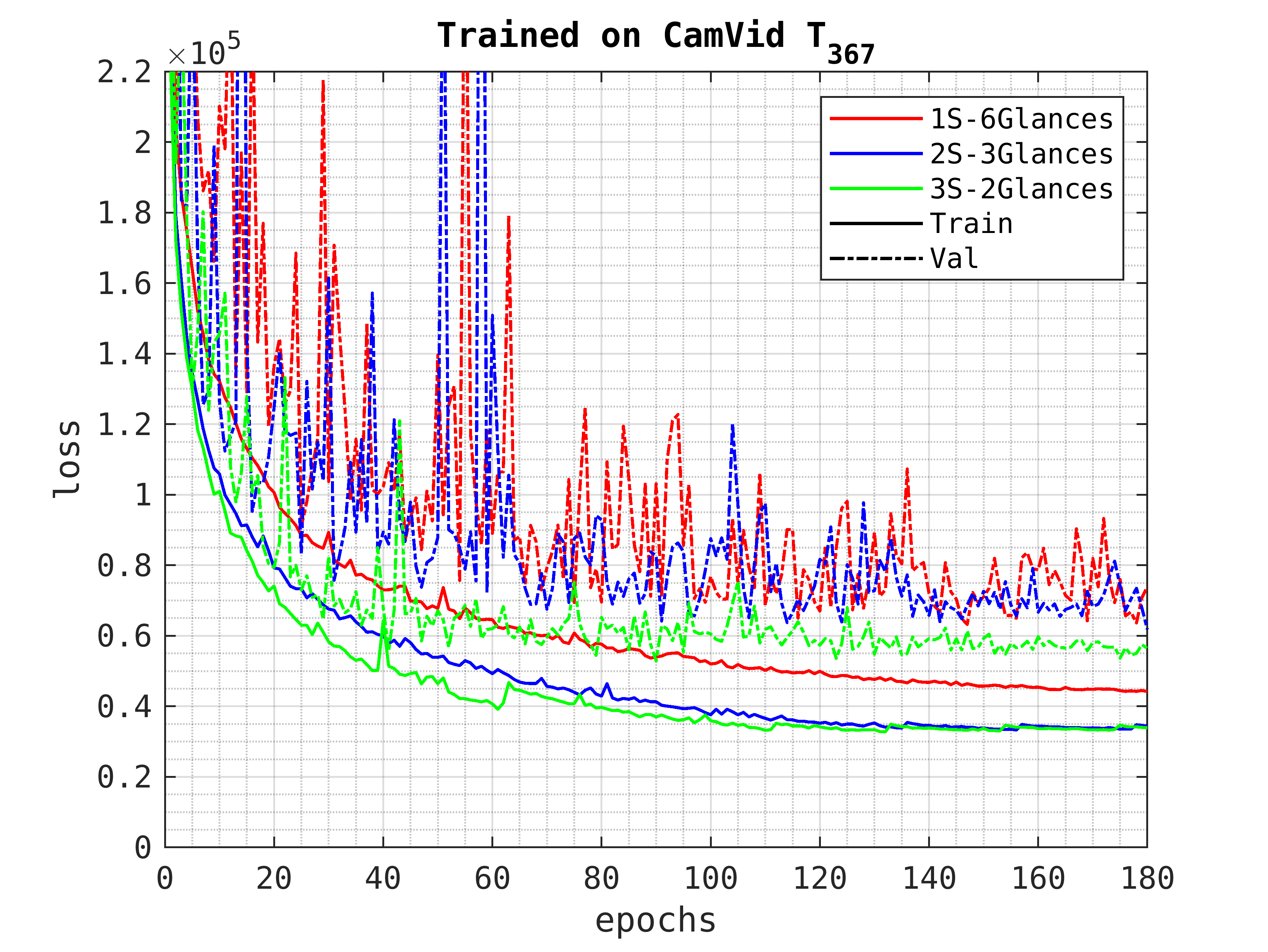}}
\subfigure{\includegraphics[width=65mm]{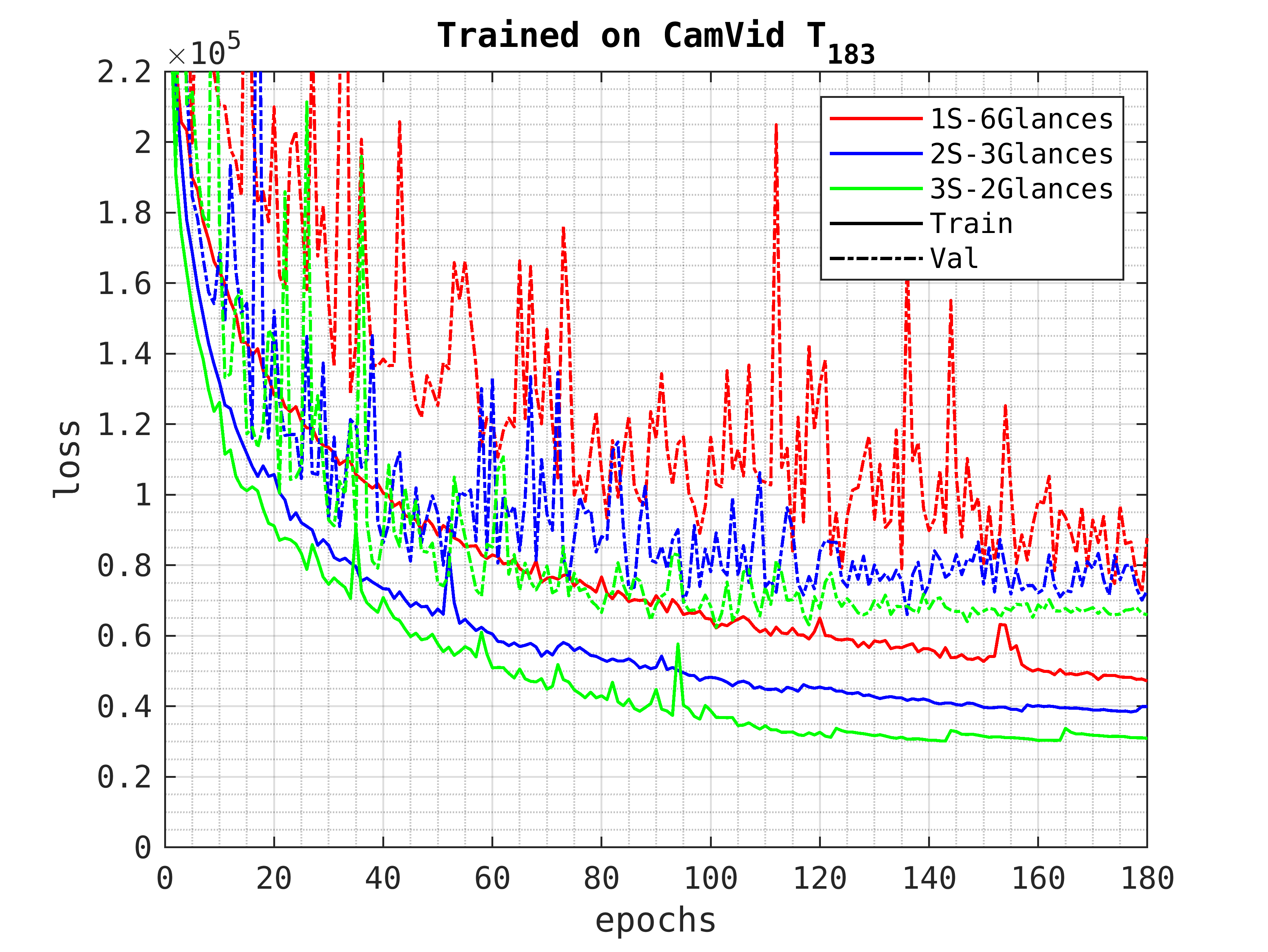}}
\subfigure{\includegraphics[width=65mm]{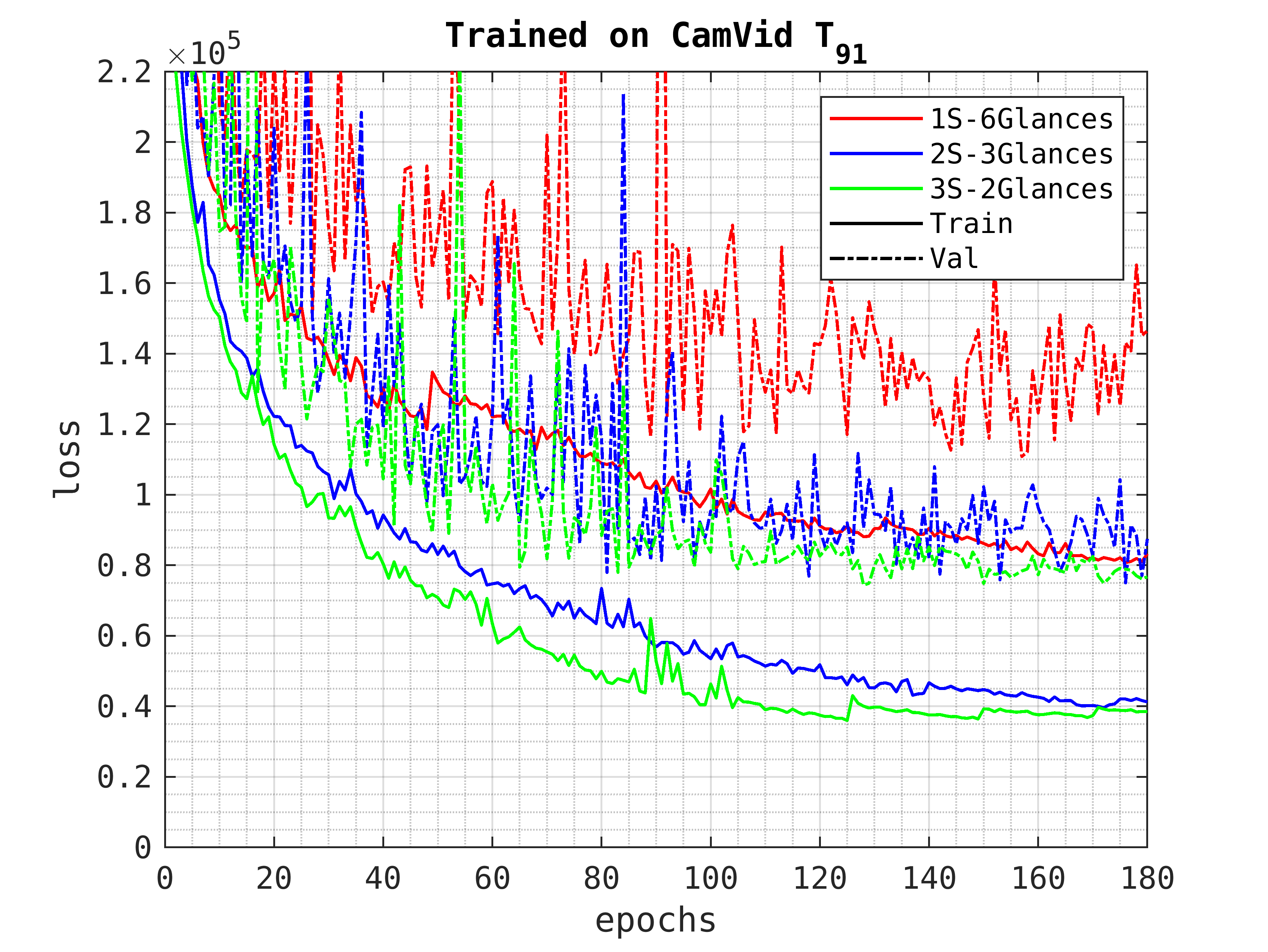}}
\caption{CamVid multi-scale ablation study training plots}
\label{C7}
\end{figure}

In Figure \ref{C7}, we present the training plots of the networks IkshanaNet-1S-6G,  IkshanaNet-2S-3G, and IkshanaNet-3S-2G, trained on three subsets of training data ($T_{367}$, $T_{183}$, and $S_{91}$). 

\begin{table}[ht]
\sisetup{detect-weight=true,detect-inline-weight=math}
\begin{center}
\begin{adjustbox}{width=1\textwidth}
\begin{tabular}{cccccccccccccccccccccc}
 \toprule
  {Subset}& {Method} & \rot{90}{road} & \rot{90}{sidewalk} &\rot{90}{building}  & \rot{90}{wall} &\rot{90}{fence}  & \rot{90}{pole} &\rot{90}{traffic light}  & \rot{90}{traffic sign} &\rot{90}{vegetation}  & \rot{90}{terrain} &\rot{90}{sky}  & \rot{90}{person} &\rot{90}{rider}  & \rot{90}{car} &\rot{90}{truck}  & \rot{90}{bus} &\rot{90}{train}  & \rot{90}{motorcycle} &\rot{90}{bicycle}  & \rot{90}{Average}      \\
  \midrule
   &ResNet18& 93.7 & 64.4 & 81.5& 14.5 & 13.8 & 27.8 & 17.8 & 26.3 & 85.0 & \bfseries 46.2 & 88.8 &  46.7 &  \bfseries 7.4 & 81.3 & \bfseries 23.8 & \bfseries 34.5 & 10.1 & 5.0 & 39.8 & 42.6\\
  &MobileNetV2& 93.9 & 64.6 & 81.8& 15.8 & 16.2 & 24.0 & 1.0 &17.6 & 84.4 & 39.9 & 88.6 & 39.2 & 0.0 & 82.5 & 13.0 & 26.8 & 9.0 & 0.0 & 32.6 & 38.5\\
   &EfficientNetb1& 93.3 & 64.9 & 81.7& 0.3 & 15.0 & 26.2 & 2.5 & 23.8 & 83.7 & 41.4 & 88.6 & 42.3 & 0.0 & 81.0 & 0.0 & 25.7 &  17.0 &0.0 &30.7 & 37.8\\
   &RegNetY08&  \bfseries 94.1 & 57.7 & 77.6& 0.0 & 0.3 & 0.0 & 0.0 & 0.0 & 82.0 & 41.4 & 88.4 & 25.7 & 0.0 & 75.0 &0.0 & 0.0 & 0.0 & 0.0 & 0.0 & 28.5\\
   $T_{1487}$ & ResNet101-1& 93.0 & 54.0 & 75.9& 12.9 & 0.9 & 3.4 & 0.0 & 0.2 & 81.7 & 38.8 & 86.9 & 30.9 & 0.0 & 73.7 & 2.7 & 0.0 & 0.0 & 0.0 & 1.0 & 29.3\\
   &DenseNet161-1& 93.1 & 57.7 & 78.9& 11.2 & 6.6 & 11.1 & 0.0 & 12.4 & 81.5 & 38.3 & 84.9 & 35.0 & 0.0 & 76.7 & 13.4 & 0.3 & 2.3 & 0.1 & 28.3 & 33.3\\
   &HRNet-V2& 92.3 & 52.3 & 76.9 &  0.7 & 0.5 & 0.0 & 0.0 & 0.0 & 81.8 & 34.5 & 82.9 & 9.4 & 0.0 & 72.3 & 0.0 & 5.3 & 0.0 & 0.0 & 18.6 & 27.8\\
   &U-Net& 94.0 & \bfseries 65.6 & \bfseries 83.2 &  13.9 & 20.5 & 34.3 & 21.7 & 43.9 & \bfseries 87.4 & 43.1 & \bfseries 89.5 & 49.9 & 0.0 & \bfseries 84.2 & 12.1 & 9.9 & \bfseries 12.9 & 0.0 & 47.9 & 42.8\\
   &IkshanaNet& 93.5 & 64.7 & 82.9& \bfseries 17.0 & \bfseries 22.6 & \bfseries 35.2 & \bfseries 27.0 & \bfseries 44.1 &  86.8 & 41.4 & 87.2 & \bfseries 52.8 & 2.5 & 81.5 & 0.3 & 25.6 & 3.9 & \bfseries 7.1 & \bfseries 48.9 & \bfseries 43.4\\

    \midrule
   &ResNet18& 92.8 & 56.3 & 78.0&  \bfseries 15.3 & 8.0 & 15.4 & 4.9 & 18.9 & 82.3 & 42.8 & 85.6 & 35.4 & 0.1 & 75.2 & \bfseries  13.1 & 13.9 & 1.9 & 0.0 & 36.6 & 35.6\\
   &MobileNetV2& 92.7 & 57.3 & 77.8& 6.1 & 7.8 & 0.7 & 0.1 & 11.3 & 81.5 & 39.2 & 85.2 & 30.8 & 0.1 & 75.9 & 3.5 &  \bfseries 22.9 & \bfseries 2.9 & 0.0 & 15.4 & 32.2\\
   &EfficientNetb1& 93.5 & 60.5 & 77.1&4.1 & 3.9 & 9.1 & 0.0 & 14.0 & 81.8 & 39.6 & 84.6 & 22.8 &  \bfseries 1.6 & 75.1 & 9.8 & 18.6 & 0.0 & 0.0 & 20.8 & 32.5\\
   &RegNetY08& \bfseries 93.9 & 58.7 & 78.8& 3.4 & 9.1 & 0.0 & 0.0 & 17.5 & 83.1 &  \bfseries 45.2 & 87.1 & 32.0 & 0.0 & 76.8 & 2.3 & 0.3 & 0.0 & 0.0 & 17.1 & 31.9\\
   $T_{743}$ &ResNet101& 90.5 & 44.6 & 72.2& 9.2 & 3.3 & 5.2 & 0.0 & 12.5 & 79.8 & 36.2 & 79.8 & 25.6 & 0.0 & 65.3 & 0.0 & 0.0 & 0.0 & 0.0 & 0.0 & 28.8\\
   &DenseNet161& 91.1 & 50.8 & 74.7& 13.9 & 3.3 & 4.4 & 1.1 & 12.1 & 78.4 & 32.1 & 80.8 & 28.6 & 0.0 & 69.6 & 2.2 & 1.4 & 0.2 & \bfseries 2.3 & 25.7 &30.1\\
   &HRNet-V2& 86.0 & 23.6 & 62.0 &  0.0 & 0.0 & 0.0 & 0.0 & 0.0 & 60.3 & 10.9 & 78.1 & 0.0 & 0.0 & 37.3 & 0.0 & 0.0 & 0.0 & 0.0 & 0.0 & 18.8\\
   &U-Net& 93.6 & \bfseries 62.1 & 81.3&  5.2 & 11.6 & 19.3 & 0.4 & 28.1 & 86.0 & 42.3 & \bfseries 87.8 & 36.5 & 0.0 & 79.6 & 9.6 & 0.5 & 0.0 & 0.0 & 5.5 & 34.2\\
   &IkshanaNet& 93.0 & 60.3 & \bfseries 81.4& 13.4 &\bfseries 23.0 & \bfseries 36.5 & \bfseries 21.9 & \bfseries 42.8 & \bfseries 86.3 & 35.1 & \bfseries 87.8 & \bfseries 42.5 & 0.0 & \bfseries 80.0 & 0.1 & 20.2 & 0.1 & 0.7 & \bfseries 38.4 & \bfseries 40.2 \\
   
   \midrule
  &ResNet18& 88.4 & 46.5 & 73.5& 2.6 & 1.1 & 4.0 & 0.0 & 5.3 & 78.4 & 36.4 & 82.9 & 27.9 & 0.0 & 68.5 & 0.0 & 0.0 & 0.0 &  \bfseries 1.4 & 13.1 & 27.9\\
   &MobileNetV2& 91.4 & 50.2 & 75.2& 10.1 &  7.3 & 5.6 & 0.3 & 8.5 & 81.0 & 34.6 & 83.0 & 27.8 & 0.0 & 72.7 &  \bfseries 16.0 &  \bfseries 1.4 & 0.0 & 0.0 & 16.6 & 30.6\\
   &EfficientNetb1& 89.8 & 46.4 & 72.4& 5.3 & 4.9 & 7.6 & 0.3 & 6.5 & 75.7 & 32.7 & 77.3 & 20.7 & 0.0 & 65.1 & 0.0 & 0.0 & 0.0 & 0.0 & 6.8 & 26.9\\
   &RegNetY08&  91.6 & 50.5 & 76.3 &  \bfseries 10.3 & 5.8 & 0.0 & 0.3 & 10.3 & 81.2 &  \bfseries 38.3 &  84.5 & 26.2 & 0.0 & 70.6 & 8.1 & 0.0 & 0.0 & 0.0 & 3.8 & 29.4\\
 $T_{371}$   &ResNet101& 89.2 & 45.8 & 73.6 & 9.8 & 2.7 & 3.1 & 0.0 & 8.0 & 79.9 & 34.8 & 81.6 & 26.9 & 0.0 & 65.3 & 8.9 & 0.0 & 0.0 & 0.0 & 13.8 & 28.6\\
   &DenseNet161& 89.0 & 42.4 & 71.9& 8.1 & 2.1 & 0.0 & 0.0 & 0.3 & 75.7 & 33.8 & 77.4 & 18.6 & 0.0 & 64.3 & 3.3 & 0.2 & 0.0 & 0.0 & 6.2 & 26.0\\
   &HRNet-V2& 87.4 & 35.3 & 68.7 &  0.1 & 0.0 & 0.0 & 0.0 & 0.0 & 74.7 & 33.9 & 76.4 & 8.9 & 0.0 & 56.2 & 0.0 & 0.0 & 0.0 & 0.0 & 0.3 & 23.3 \\
   &U-Net&\bfseries 92.3 & \bfseries 57.4 & 76.9 &  4.7 & \bfseries 7.4 & 2.7 & \bfseries 1.2 & 14.1 & 83.4 & 36.6 & \bfseries 85.8 & 28.5 & 0.0 & \bfseries 75.8  & 3.3 & 0.0 & \bfseries 1.9 & 0.0 & 1.4 & 30.2\\
   &IkshanaNet& 90.5 & 51.6 & \bfseries 78.1& 4.5 & 1.7 & \bfseries 28.0 &  0.7 & \bfseries 24.0 & \bfseries 84.0 & 33.4 & 83.9 & \bfseries 35.2 & 0.0 & 65.5 & 0.1 & 0.8 & 0.0 & 0.0 & \bfseries 20.3 & \bfseries 31.7\\
   \midrule
   &ResNet18& 85.0 & 32.3 & 70.4& 1.6 & 0.0 & 0.0 & 0.0 & 0.0 & 75.9 & 32.0 & 75.2 & 0.0 & 0.0 & 53.0 & 0.0 & 0.0 & 0.0 & 0.0 & 0.0 & 22.4\\
  &MobileNetV2& 85.9 & 37.3 & 68.8& 0.0 & 0.0 & 0.0 & 0.0 & 0.0 & 74.1 & \bfseries 34.6 & 71.3 & 0.0 & 0.0 & 54.6 & 0.0 & 0.0 & 0.0 & 0.0 & 0.0 & 22.5\\
   &EfficientNetb1& 85.2 & 38.8 & 71.0& 4.1 & 1.5 & 0.0 & 0.0 & 0.2 & 78.6 & 31.8 & 80.0 & 17.9 & 0.0 & 58.1 &0.0 & 0.0 & 0.0 & 0.0 & 0.0 & 24.6\\
   &RegNetY08& 89.4 & 47.2 & 74.3&  6.5 & 2.5 & 0.0 & 0.2 & 8.2 & 78.9 &  35.7 &  82.6 & 22.4 & 0.0 &  \bfseries 66.9 & 3.1 & 0.0 & 0.0 & 0.0 & 2.2 & 27.4\\
     $T_{185}$&ResNet101& 85.4 & 29.7 & 66.9& 0.0 & 0.0 & 0.0 & 0.0 & 0.0 & 73.4 & 31.3 & 75.0 & 0.0 & 0.0 & 48.5 & 0.0 & 0.0 & 0.0 & 0.0 & 0.0 & 21.6\\
   &DenseNet161& 87.8 & 40.2 & 72.2& 3.2 & 2.4 & 0.0 & 0.0 & 0.0 & 76.9 & 28.3 & 78.6 & 15.2 & 0.0 & 61.5 & 2.3 & 0.0 & 0.0 & 0.0 & 5.1 & 24.9\\
   &HRNet-V2& 86.4 & 19.3 & 63.8 &  0.0 & 0.0 & 0.0 & 0.0 & 0.0 & 67.9 & 0.0 & 70.6 & 0.0 & 0.0 & 38.8 & 0.0 & 0.0 & 0.0 & 0.0 & 0.0 & 18.3\\
   &U-Net& \bfseries 90.2 & \bfseries 50.1 & 74.0 &  \bfseries 7.5 & 0.6 & 0.0 & 0.0 & 8.1 & 81.1 & 33.1 & \bfseries 82.7 & 13.2 & 0.0 & 68.4 & 0.0 & 0.0 & \bfseries 0.6 & \bfseries 0.2 & \bfseries 18.4 & 27.8\\
   &IkshanaNet&  90.1 & 49.2 & \bfseries 74.6 & 5.9 & \bfseries 3.8 & \bfseries 17.7 & \bfseries 2.0 & \bfseries 18.7 & \bfseries 81.6 & 33.6 & 80.7 & \bfseries 29.6 & 0.0 & \bfseries66.9 & \bfseries 3.2 & 0.0 & 0.0 & 0.0 & 10.2 & \bfseries 29.9\\
   
   \midrule
   &ResNet18& 85.7 & 26.4 & 66.4& 0.0 & 0.0 & 0.0 & 0.0 & 0.0 & 74.2 & 26.9 & 72.9 & 0.0 & 0.0 & 47.4 & 0.0 & 0.0 & 0.0 & 0.0 & 0.0 & 21.0\\
   &MobileNetV2& 82.1 & 20.4 & 63.7& 0.0 & 0.0 & 0.0 & 0.0 & 0.0 & 69.2 & 25.5 & 67.7 & 0.0 & 0.0 & 36.6 & 0.0 & 0.0 & 0.0 & 0.0 & 0.0 & 19.2\\
   &EfficientNetb1& 84.7 & 26.6 & 59.8& 1.6 & 0.0 & 0.0 & 0.0 & 0.1 & 67.2 & 29.9 & 58.6 &2.4 & 0.0 & 46.1 & 0.0 & 0.0 & 0.0 & 0.0 & 0.0 & 19.8\\
   &RegNetY08& 86.7 & 33.4 & 68.5 & 0.0 & 0.0 & 0.0 & 0.0 & 0.0 & 74.8 & 27.9 & 76.3 & 0.0 & 0.0 & 52.8 & 0.0 & 0.0 & 0.0 & 0.0 & 0.0 & 22.1\\
  $T_{92}$ &ResNet101& 81.4 & 20.7 & 62.2 & 0.0 & 0.0 & 0.0 & 0.0 & 0.0 & 69.4 & 28.2 & 68.6 & 0.0 & 0.0 & 39.9 & 0.0 & 0.0 & 0.0 & 0.0 & 0.0 & 19.4\\
  &DenseNet161& 84.0 & 24.3 & 67.5 & 0.0 & 0.3 & 0.0 & 0.0 & 0.0 & 72.7 & 23.0 & 74.6 & 0.0 & 0.0 & 49.3 & 0.0 & 0.0 & 0.0 & 0.0 & 0.0 & 20.8\\
   &HRNet-V2& 81.0 & 3.2 & 49.5 &  0.0 & 0.0 & 0.0 & 0.0 & 0.0 & 52.0 & 0.0 & 67.2 & 0.0 & 0.0 & 38.9 &  0.0 & 0.0 & 0.0 & 0.0 & 0.0& 15.4\\
   &U-Net& \bfseries 87.6 & \bfseries 45.8 & 71.3 &  2.2  & 0.0 & 0.0 & 0.0 & \bfseries 5.8 & 79.2 & 31.1 & \bfseries 79.8 & 14.2 & 0.0 & 58.4 & 0.0 & 0.0 & \bfseries 0.3 & 0.0 & 0.0 & 25.0 \\
   &IkshanaNet&  87.0 &  40.0 & \bfseries 72.3 & \bfseries 3.0 & \bfseries 3.2 & \bfseries 10.4 & 0.0 & 2.3 & \bfseries 79.3 & \bfseries 31.3 & 79.1 & \bfseries 20.8 & 0.0 & \bfseries 58.9 & 0.0 & 0.0 & 0.0 & 0.0 & \bfseries 2.9 & \bfseries25.8\\
   \bottomrule
\end{tabular}
\end{adjustbox}
\end{center}
 \caption{Class-wise results of the Cityscapes data ablation study evaluated on val set}
 \label{a-table-2}
\end{table}

\begin{table}[ht]
\sisetup{detect-weight=true,detect-inline-weight=math}
\begin{center}
\begin{adjustbox}{width=1\textwidth}
\begin{tabular}{llSSSSSSSSSSSSSSSSSSSS}
 \toprule
   {Subset}&{Method} & \rot{90}{road} & \rot{90}{sidewalk} &\rot{90}{building}  & \rot{90}{wall} &\rot{90}{fence}  & \rot{90}{pole} &\rot{90}{traffic light}  & \rot{90}{traffic sign} &\rot{90}{vegetation}  & \rot{90}{terrain} &\rot{90}{sky}  & \rot{90}{person} &\rot{90}{rider}  & \rot{90}{car} &\rot{90}{truck}  & \rot{90}{bus} &\rot{90}{train}  & \rot{90}{motorcycle} &\rot{90}{bicycle}  & \rot{90}{Average}      \\
  \midrule
   &1S-6Glances& 86.0 & 44.5 & 72.2 & 4.0 & 3.3 & 21.9 & 2.0 & 28.0 & 83.1 & 35.4 & 77.5 & 29.7 & 0.0 & 52.8 & 0.0 & 0.9 & 0.0 & 0.0 & 12.7 & 29.2 \\
   $S_{1487}$&2S-3Glances& 90.9 & 57.5 & 78.6& 9.0 & 13.8 & 34.6 & 16.8 & 42.8 & 85.3 & 39.9 & 83.7 & 42.2 & 0.0 & 73.1 & 0.0 & 5.4 & 0.2 & 0.0 & 35.0 & 37.3\\
   &3S-2Glances& \bfseries 94.0 & \bfseries 67.6 & \bfseries 83.0 & \bfseries 16.4 & \bfseries 24.7 & \bfseries 39.1 & \bfseries 23.7 & \bfseries 47.1 & \bfseries 87.0 & \bfseries 41.7 & \bfseries 87.9 & \bfseries 50.7 & \bfseries 2.2 & \bfseries 81.1 & \bfseries 3.6 & \bfseries 18.7 & \bfseries 5.4 & \bfseries 2.8 & \bfseries 49.6 & \bfseries 43.5\\
    \midrule
   &1S-6Glances& 85.0 & 34.0 & 67.6& 0.2 & 0.3 & 17.1 & 0.6 & 15.4 & 80.4 & 30.0 & 74.8 & 20.9 & 0.0 & 45.2 & 0.0 & 0.0 & 0.0 & 0.0 & 1.4 & 24.9\\
   $S_{743}$&2S-3Glances& 90.8 & 56.8 & 77.4 & 8.6 & 13.3 & \bfseries 31.6 & \bfseries 12.6 & \bfseries 33.6 & \bfseries 85.8 & 38.1 & 84.1 & 38.3 & 0.0 & 66.9 & 0.1 & 0.0 & 0.0 & 0.1 & 24.3 & 34.9\\
   &3S-2Glances& \bfseries 92.3 & \bfseries 60.5 & \bfseries 80.0 & \bfseries 11.2 & \bfseries 14.7 & \bfseries 31.6 & 6.2 & \bfseries 33.6 & 85.5 & \bfseries 39.7 & \bfseries 86.8 & \bfseries 41.4 & 0.0 & \bfseries 73.9 & \bfseries 4.6 & \bfseries 2.3 & \bfseries 6.8 & \bfseries 2.1 & \bfseries 28.8 & \bfseries 36.9\\
   \midrule
   &1S-6Glances& 80.6 & 29.4 & 65.1& 1.0 & 0.3 & 8.8 & 1.5 & 8.8 & 79.0 & 27.7 & 73.8 & 22.3 & 0.0 & 43.9& 0.0 & 0.0 & 0.0 & 0.0 & 1.1 & 23.0\\
   $S_{371}$&2S-3Glances& 88.6 & 51.4 & 74.3& 5.9 & 10.4 & 26.1 & 0.9 & \bfseries 28.3 & \bfseries 83.8 & \bfseries 36.5 & 82.7 & 37.3 & 0.0 & 66.7 & 0.6 & 
   \bfseries 6.3 & \bfseries 2.4 & 0.0 & \bfseries 28.8 & 33.2 \\
   &3S-2Glances& \bfseries 90.7 & \bfseries 56.0 & \bfseries 78.1& \bfseries 12.0 & \bfseries 12.0 & \bfseries 26.8 & \bfseries 4.5 & 26.1 & 83.6 & 33.6 & \bfseries 85.4 & \bfseries 38.5 & 0.0 & \bfseries 70.7 & \bfseries 3.8 & 3.5 & 0.0 & \bfseries 0.2 & 27.6 & \bfseries 34.4\\
   \midrule
   &1S-6Glances& 80.0 & 18.2 & 59.1& 0.0 & 0.0 & 6.0 & \bfseries 0.2 & 2.9 & 73.3 & 26.0 & 72.0 & 8.6 & 0.0 & 35.7 & 0.8 & 0.0 & 0.0 & 0.0 & 0.0 & 20.2\\
   $S_{185}$&2S-3Glances& 84.8 & 39.5 & 69.3& \bfseries 3.6 & 0.4 & 7.8 & 0.0 & \bfseries 12.8 & 78.8 & \bfseries 32.7 & 79.9 & 19.3 & 0.0 & 52.0 & 0.0 & 0.0 & 0.0 & 0.0 & \bfseries 8.0 & 25.7\\
   &3S-2Glances& \bfseries 88.7 & \bfseries 43.4 & \bfseries 72.9 & 1.5 & \bfseries 3.7 & \bfseries 12.9 & \bfseries 0.2 & 12.7 & \bfseries 81.1 & 31.5 & \bfseries 80.0 & \bfseries 24.9 & 0.0 & \bfseries 61.6 & \bfseries 2.2 & 0.0 & 0.0 & 0.0 & 4.4 & \bfseries 27.5\\
   \midrule
   &1S-6Glances& 77.7 & 11.1 & 52.5& 0.0 & 0.0 & 0.0 & \bfseries 0.1 & 3.6 & 71.7 & 25.0 & 70.5 & 1.9 & 0.0 & 30.5 & 0.0 & 0.0 & 0.0 & 0.0 & 0.0 & 18.1\\
   $S_{92}$&2S-3Glances& 82.7 & 32.9 & 66.5& 2.4 & 0.1 & \bfseries 8.7 & 0.0 & \bfseries 10.9 & 78.0 & 29.1 & 76.7 & 18.8 & 0.0 & 46.8 & 0.0 & 0.0 & \bfseries 0.1 & 0.0 & 2.0 & 24.0\\
   &3S-2Glances& \bfseries 86.8 & \bfseries 34.6 & \bfseries 71.6 & \bfseries 4.5 & \bfseries 6.6 & 8.3 & 0.0 & 10.8 & \bfseries 79.5 & \bfseries 32.7 & \bfseries 78.0 & \bfseries 24.7 & 0.0 & \bfseries 59.5 & \bfseries 0.3 & 0.0 & 0.0 & 0.0 & \bfseries 5.0 & \bfseries 26.5\\
   \bottomrule
\end{tabular}
\end{adjustbox}
\end{center}
\caption{Class-wise IoU results of the Cityscapes multi-scale ablation study evaluated on val set}
 \label{a-table-3}
\end{table}

\end{document}